\documentclass[11pt]{article}
\usepackage[preprint]{acl}

\usepackage{times}
\usepackage{latexsym}
\usepackage{todonotes}

\usepackage{colortbl}
\definecolor{cmigblue}{RGB}{219,238,247}
\newcommand{\cmigstrut}{\rule[-0.45em]{0pt}{1.56em}}
\usepackage{arydshln}

\usepackage{booktabs} % 用于绘制专业的三线表
\usepackage{multirow} % 用于跨行合并单元格

\newlength{\tblsep}
\usepackage{graphicx} % 用于\resizebox命令缩放表格
\usepackage{adjustbox}  % 表格缩放（若需要）
\usepackage{rotating}   % 横向表格（可选）
\usepackage{placeins}   % \FloatBarrier

\usepackage{amsmath}     % 数学环境
\usepackage{geometry}
\usepackage[most]{tcolorbox}
\usepackage{xcolor}
\usepackage{listings}
\tcbset{
  promptbox/.style={
    enhanced, breakable, colback=gray!5, colframe=black!70,
    boxrule=0.6pt, arc=2pt, left=6pt, right=6pt, top=4pt, bottom=4pt,
    fonttitle=\bfseries\small, coltitle=white, colbacktitle=black!70,
    attach boxed title to top left={xshift=4pt, yshift=-2pt},
    boxed title style={colback=black!70, sharp corners},
    fontupper=\small\ttfamily,
  }
}

\usepackage[T1]{fontenc}
\usepackage[utf8]{inputenc}

\usepackage{microtype}

\usepackage{inconsolata}

\usepackage{graphicx}
\usepackage{amsmath}
\usepackage{amssymb}
\usepackage{booktabs}

\title{C‑MIG: Multi‑view Information Gain-based Retrieval‑Augmented Generation for Clinical Diagnosis Reasoning}

\author{
  \textbf{Yuwei Miao\textsuperscript{1,*}},
  \textbf{Yunsheng Zeng\textsuperscript{1,*}},
  \textbf{Gen Li\textsuperscript{1}},
  \textbf{Xiandong Li\textsuperscript{1}},
  \textbf{Yujin Wang\textsuperscript{1}},
  \textbf{Siyu Chen\textsuperscript{1}}
  \\
  \textbf{Luning Wang\textsuperscript{1}},
  \textbf{Yunhao Qiao\textsuperscript{1}},
  \textbf{Junfeng Wang\textsuperscript{1}},
  \textbf{Jianwei Lv\textsuperscript{1,\dag}},
  \textbf{Bo Yuan\textsuperscript{1,\dag}}
  \\[2pt]
  \textsuperscript{1}Baidu Inc.
}

\begin{document}
\maketitle

\renewcommand{\thefootnote}{\fnsymbol{footnote}}
\footnotetext[1]{Equal contribution.}
\footnotetext[2]{Corresponding author. E-mail: \{yuanbo07,lvjianwei\}@baidu.com}
\renewcommand{\thefootnote}{\arabic{footnote}}

\begin{abstract}

Retrieval-augmented generation (RAG) trained with reinforcement learning (RL) is a promising way to ground large language models in medical evidence for clinical diagnosis. However, existing RAG-RL methods face two problems: (1) their process reward is largely single-view, as most methods inherited from open-domain multi-hop QA reward only the quality of the retrieved information; (2) assigning a reliable process reward is hard, as hard matching gives zero to clinically correct but non-verbatim steps such as ``Allergic Sinusitis'' versus ``Allergic Rhinitis'', and the alternative of strong-teacher guiding is bounded by the teacher's own medical competence. To address these issues, we propose \textbf{C-MIG}, a \textbf{M}ulti-view \textbf{I}nformation \textbf{G}ain-based framework for \textbf{C}linical diagnosis. We identify two complementary and crucial intermediate steps in clinical diagnosis: \textit{what to retrieve} and \textit{how to refine}. We reward both by the information gain each brings, measured as the increase in the predicted likelihood of the ground-truth answer under a base, frozen reference model. The frozen scorer yields a stable and reproducible reward that needs no extra reward model or teacher. Experiments on four medical benchmarks show that C-MIG outperforms RAG-RL baselines in and out of domain and stays competitive with far larger medical-specialist and general-purpose LLMs, at only about 3.4\% training overhead and no inference-time cost.
% Comprehensive experiments on four medical benchmarks, C-MIG surpasses IGPO by 6.71 points in-domain and achieves the best out-of-domain performance, offering a reliable and generalizable RAG-RL solution for clinical diagnosis.

\end{abstract}

\section{Introduction}
Large language models (LLMs) have been increasingly applied to clinical diagnosis~\citep{dou2026baichuan,qiu2025quantifying,liao2025ehr}, aiming to provide efficient and accurate responses to patient queries.
However, outdated knowledge, insufficient domain specialization, and the limitations of parametric memorization cause LLMs to hallucinate~\citep{orgad2025llms,huang2025survey,feng2026hyper}, posing direct risks to patient safety.

\begin{figure}[!t]
    \centering
    \includegraphics[width=\columnwidth]{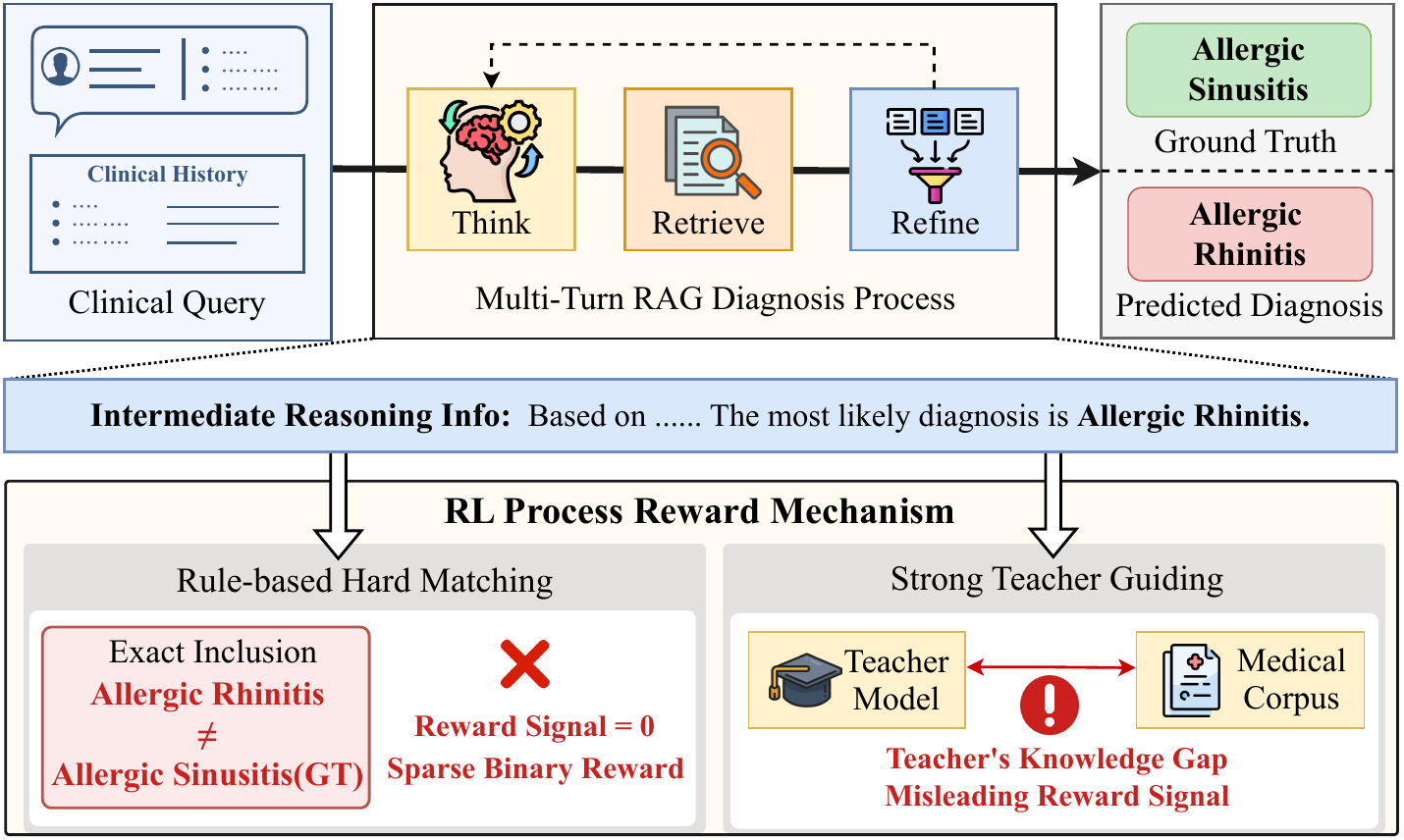}
    \caption{Both common process-reward schemes fail in clinical diagnosis. Rule-based hard matching gives zero reward to semantically relevant but non-verbatim steps, yielding sparse signals; strong-teacher guiding is capped by the teacher's medical competence and propagates its knowledge gaps.
    }
    \label{fig:introduce}
\end{figure}

Retrieval-augmented generation (RAG) ~\citep{li2025enhancing,wang2026chain} has emerged as a mainstream solution to mitigate LLM hallucinations by dynamically integrating external knowledge into the generation process, improving factual reliability. Recent studies~\citep{jin2025search,song2025r1,wang2026information,shi2026search,wang2026prorag} further draw inspiration from reinforcement learning (RL) and bring RAG into RL-based reasoning. By rewarding only the correctness of the final answer, they teach the model to autonomously issue multiple retrieval turns and weave external medical evidence into its reasoning chain, already achieving promising results.

However, real-world clinical reasoning unfolds over multiple intermediate steps. As the mainstream RL training method, GRPO~\citep{guo2025deepseek} rewards only the final answer and leaves these steps without fine-grained supervision. This calls for a process reward over the intermediate steps; yet existing process-reward assignment methods face the following challenges:
\begin{itemize}
    \item \textbf{Single-view process reward.} Existing RAG-RL methods~\citep{wang2026information} are largely built for open-domain multi-hop QA and reward only the quality of retrieved information. In medicine the retrieved knowledge must still be internally understood and transformed before it can support a diagnosis. A rigorous diagnosis therefore depends not only on \emph{what is retrieved} but on whether the model can distill the truly diagnostic evidence from it.
    \item \textbf{Hard process-reward assignment.} When a process reward is desired, assigning it reliably is hard, and as shown in Figure~\ref{fig:introduce}, the two common remedies both fail in medicine. Rule-based hard matching~\citep{shi2026search} assigns zero to clinically relevant but non-verbatim steps such as ``Allergic Sinusitis'' versus ``Allergic Rhinitis'', yielding sparse rewards. The alternative lets a strong teacher guide the intermediate steps through rubric judging~\citep{gunjal2026rubrics} or on-policy distillation~\citep{agarwal2024policy}, which caps reliability at the teacher's medical competence and requires substantial additional training overhead.
\end{itemize}

To address these challenges, we propose \textbf{C-MIG}, a \textbf{M}ulti-view \textbf{I}nformation \textbf{G}ain-based retrieval-augmented generation framework for \textbf{C}linical diagnosis. For the first problem, we build on the clinical-reasoning principle that a diagnosis is reached by first gathering evidence and then synthesizing it~\citep{sackett1996evidence,schaller2021lessons}, and we supervise these two complementary steps separately. The first step is \emph{what to retrieve}, where query generation poses high-quality queries that narrow the differential diagnosis; the second is \emph{how to refine}, where evidence refinement distills from lengthy and noisy documents the findings that confirm or exclude candidate diseases. Prior methods reward only one of these two steps and most often retrieval alone, whereas clinical diagnosis requires both. For the second problem, we argue that the medical knowledge used during RAG-RL should reside in a comprehensive and updatable medical corpus rather than be internalized from a teacher model. Inspired by methods that use the model's own predicted likelihood to judge whether an intermediate step is beneficial~\citep{wang2026information,ma2025proof}, we reward both retrieval and refinement by the information gain each brings toward the ground-truth answer. This solves the hard-matching problem and does not depend on the medical competence of the teacher model, since judging whether information increases the likelihood of a given answer is a basic capability. Unlike existing methods that score with the continuously updated policy model, the heavily overlapping terminology of clinical diagnosis requires a stable scorer rather than a shifting criterion, so we use a base, frozen reference model that gives a reproducible process reward and runs only during training.

In summary, our contributions are as follows:
\begin{itemize}
    \item \textbf{Dual-view process reward grounded in clinical reasoning.} Starting from the cognitive nature of clinical diagnosis, we explicitly decouple the RAG trajectory into two complementary steps, \emph{what to retrieve} and \emph{how to refine}, and supervise each separately. This addresses prior designs that reward only one of these steps, most often retrieval quality alone.
    \item \textbf{A more reliable process reward.} Using a base, frozen reference model, we adopt the likelihood gain of the ground-truth answer from each retrieval and refinement step as the process reward. This yields a continuous reward over intermediate steps, addressing both the sparsity of 0/1 hard matching on semantically similar medical terms and the dependence on an expensive, unreliable medical-specialist teacher.
    \item \textbf{Extensive empirical validation.} We demonstrate through extensive experiments on four medical benchmarks that C-MIG achieves the best overall average among RAG-RL methods both in and out of domain, and remains competitive with much larger medical-specialist and general-purpose LLMs. Moreover, C-MIG adds only about 3.4\% per-step training overhead and no inference-time overhead.
\end{itemize}

\section{Related Work}

\textbf{RAG and Reinforcement Learning.}
Early RAG systems employ single-round retrieval with inherent limitations~\citep{xiong2024benchmarking}. Recent works integrate RAG with RL for multi-round retrieval and reasoning~\citep{li2026ur2,song2025r1,shi2026search}: ReSearch~\citep{chen2025research} embeds search into reasoning chains, s3~\citep{jiang2025s3} decouples the searcher from the generator, MMOA-RAG~\citep{chen2026improving} formulates RAG modules as cooperative multi-agents, and DynamicRAG~\citep{sun2026dynamicrag} trains a reranker with generation-quality feedback. These methods either optimize a single module or lack fine-grained credit assignment over intermediate steps. C-MIG fills this gap with multi-view information gain over \emph{what to retrieve} and \emph{how to refine}.

\textbf{Clinical RAG.}
LLM reasoning has extended to complex clinical diagnosis~\citep{ma2026medla,peng2025tree,xia2026mmedagent}. MedRAG~\citep{zhao2025medrag} fuses diagnostic knowledge graphs (KGs) with electronic health records (EHRs), MedGraphRAG~\citep{wu2025medical} builds medical graph structures for evidence-based generation, and RAG$^2$~\citep{sohn2025rationale} filters retrieved passages via rationale models. For medical RAG-RL, \citet{zheng2025end} trains an agentic diagnostic system with a threefold RL framework, and \citet{lu2026medr3enhancingmedicalretrievalaugmented} designs multi-dimensional medical rewards covering entity coverage and KG path alignment. Existing clinical methods rely on exact-match or KG-alignment rewards that produce zero signal for semantically relevant outputs; C-MIG provides continuous information gain rewards suited to the naming ambiguity inherent in clinical diagnosis.

\textbf{Process Reward for Multi‑Turn Reasoning.}
Outcome-only rewards suffer from poor credit assignment, motivating process rewards over intermediate steps. Rule-based hard matching rewards a step only when it verbatim contains the gold answer, giving zero signal to semantically correct but non-verbatim steps~\citep{jin2025search}. A second line relies on a strong teacher to guide: rubric judging~\citep{gunjal2026rubrics}, a process reward model trained via Monte Carlo tree search (MCTS)~\citep{wang2026prorag}, an LLM judge of document relevance~\citep{li2025r3}, on-policy distillation~\citep{agarwal2024policy}, or trained stage‑wise and hierarchical rewards~\citep{zhang2026process,wu2026hiprag}, all bounded by the teacher's own competence. A third line scores a step by the model's own likelihood of the answer~\citep{wang2026information,setlur2025rewarding}. C-MIG follows this third line but computes the likelihood with a frozen reference model, which yields a more stable reward signal in the medical domain while avoiding both the sparsity of hard matching and the capability ceiling of a teacher.

\section{Method}
\label{sec:method}

\begin{figure*}[t]
    \centering
    \includegraphics[width=\textwidth]{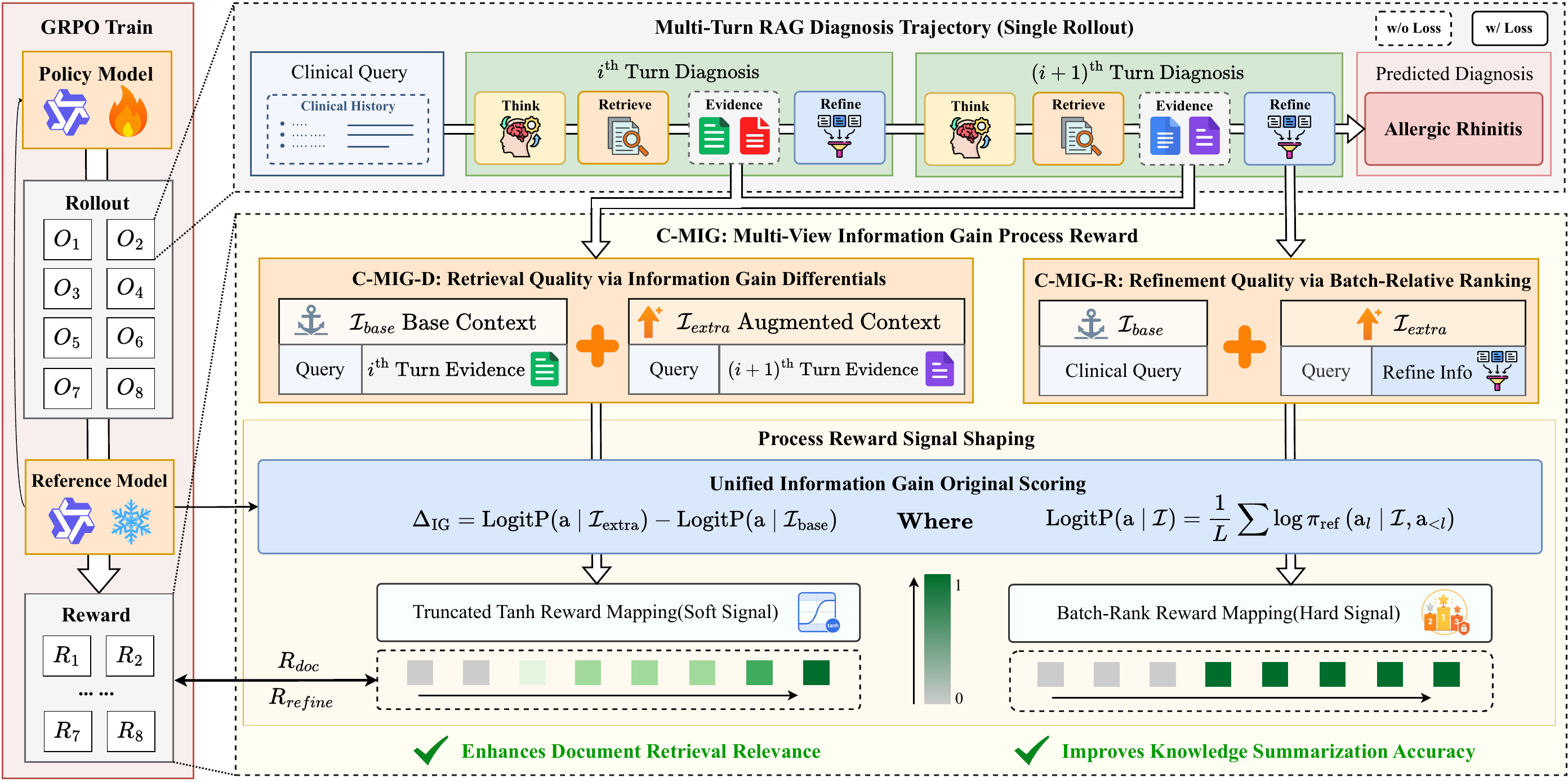}
    \caption{Overview of C-MIG. Top: the policy rolls out a multi-turn trajectory of think, search, evidence, and refine steps ending in a diagnosis, with evidence tokens masked from the loss. Bottom: a frozen reference model scores it from two views, C-MIG-D for retrieval quality via cross-turn gain differentials and C-MIG-R for refinement quality via batch-relative ranking, shaped into rewards by truncated tanh and batch-rank mappings.}
    \label{fig:model}
    \vspace{4pt}
\end{figure*}

As shown in Figure~\ref{fig:model}, C-MIG, built on the joint RAG-RL framework, uses a frozen reference model to compute information gain from two complementary views, \emph{what to retrieve} and \emph{how to refine}, yielding a fine-grained process reward over the clinical-diagnosis trajectory. This section describes C-MIG in detail.

% This section presents our training framework. We describe reasoning trajectory generation (\S\ref{sec:trajectory}), the Logit-P process reward framework (\S\ref{sec:logitp}), its instantiation as Refine-P (\S\ref{sec:refinep}) and Doc-P (\S\ref{sec:docp}), and outcome rewards with the training objective (\S\ref{sec:training}). The overall framework is illustrated in Figure~\ref{fig:model}.

\subsection{Rollout Generation}
\label{sec:trajectory}

\textbf{Task Formulation} Given a clinical query $q$ and its diagnostic label $a$, the policy model $\pi_\theta$ generates a rollout through multi-turn interaction with a retrieval engine $\mathcal{E}$. Each round comprises four structured operations: \texttt{<think>} analyzes the patient query information and plans retrieval directions, \texttt{<search>} issues a query to the retrieval engine, \texttt{<evidence>} receives the top-$k$ most relevant documents returned by $\mathcal{E}$, and \texttt{<refine>} summarizes them, extracting critical information and filtering noise. After several think-search-refine cycles, the model outputs its conclusion in \texttt{<diagnosis>} tags, where the number of cycles is determined autonomously and the generation process terminates upon producing a diagnosis or reaching the maximum allowed number of cycles $T_{\max}$. Complete prompts are provided in Appendix \ref{appendix_prompt}.

\subsection{Multi-View Information Gain}
\label{sec:logitp}

As discussed in \S1, process rewards for clinical RAG face two problems: they are often single-view, as existing methods supervise only one of the two steps and most often retrieval alone; and reliable assignment is hard, as exact-match rules give zero signal to semantically relevant but non-verbatim steps, while a strong teacher is capped by its own competence. We propose a multi-view information gain method that addresses both with a continuous signal: if the retrieval or refinement process facilitates diagnosis, the frozen reference model's predicted likelihood of the correct answer should increase, and otherwise decrease.

Formally, given a frozen base model $\pi_{\text{ref}}$ and the gold answer $a = (a_1, \ldots, a_L)$ of length $L$, we define the log-probability scoring function:
\begin{equation}
\text{LogitP}(a \mid \mathcal{I}) = \frac{1}{L} \sum_{l=1}^{L} \log \pi_{\text{ref}}(a_l \mid \mathcal{I}, a_{<l})
\end{equation}
where $\mathcal{I}$ denotes the contextual information provided to the reference model. Based on this, we define the unified information gain as:
\begin{equation}
\label{eq:unified_ig}
\Delta_{\text{IG}} = \text{LogitP}(a \mid \mathcal{I}_{\text{extra}}) - \text{LogitP}(a \mid \mathcal{I}_{\text{base}})
\end{equation}
where $\mathcal{I}_{\text{base}}$ is the base context before incorporating new information, and $\mathcal{I}_{\text{extra}}$ is the augmented context after incorporating the intermediate output, which is either the retrieved documents or the refined knowledge. $\Delta_{\text{IG}} > 0$ indicates that the new information reduces diagnostic uncertainty, while $\Delta_{\text{IG}} \leq 0$ indicates noise introduction.

\paragraph{Distribution Shift Avoidance.}
All likelihoods are computed by the frozen $\pi_{\text{ref}}$, not the continuously updated $\pi_\theta$. \citet{wang2026information} uses the policy model for this computation, suffering from reward drift as the policy distribution shifts during training. Our frozen $\pi_{\text{ref}}$ provides a time-invariant anchor, guaranteeing reward consistency throughout training.

We instantiate Eq.~\ref{eq:unified_ig} from two complementary views to address the credit assignment problem: C-MIG-D (\S\ref{sec:docp}) evaluates cross-turn retrieval quality by measuring information gain between successive rounds, and C-MIG-R (\S\ref{sec:refinep}) evaluates intra-turn refinement quality by measuring the information gain of knowledge summarization.

\subsubsection{C-MIG-D: Retrieval Quality via Information Gain Differentials}
\label{sec:docp}

C-MIG-D progressively incentivizes the model to retrieve more valuable documents by measuring cross-turn information gain. For a trajectory with $N$ rounds, we instantiate Eq.~\ref{eq:unified_ig} by setting $\mathcal{I}_{\text{base}} = (q, D_{k-1})$ and $\mathcal{I}_{\text{extra}} = (q, D_k)$, where $D_k$ denotes the documents returned in round $k$. The information gain is:
\begin{equation}
\Delta(D_k) = \text{LogitP}(a \mid q, D_k) - \text{LogitP}(a \mid q, D_{k-1})
\end{equation}
Inter-round improvement is measured by:
\begin{equation}
\delta_k = \Delta(D_k) - \Delta(D_{k-1}), \quad k = 2, \ldots, N
\end{equation}
To avoid introducing noise, we add two constraints: (1) First-Round Quality Gate: C-MIG-D reward is zero if $\Delta(D_1) \leq 0$, preventing reward hacking where the model deliberately performs ineffective first-round retrieval to inflate subsequent information gain, (2) Half-Positive Scheme, that truncates negative $\delta_k$ to zero.
The combined reward is mapped via a truncated tanh function:
\begin{equation}
\label{eq:doc-p}
R_{\text{doc}} = \begin{cases} 0, \quad\quad\quad\quad\quad\quad \Delta(D_1) \leq 0 \text{ or } N < 2 \\[4pt] \displaystyle \frac{w_d}{N} \sum_{k=2}^{N} \max\!\big(\tanh(\alpha_d \delta_k), 0\big), \text{otherwise} \end{cases}
\end{equation}
where $\alpha_d$ is a sensitivity parameter and $w_d$ is a weight coefficient. 

\subsubsection{C-MIG-R: Refinement Quality via Batch-Relative Ranking}
\label{sec:refinep}

C-MIG-R measures the actual contribution of the refined content to diagnostic confidence by measuring intra-turn information gain. We instantiate Eq.~\ref{eq:unified_ig} by setting $\mathcal{I}_{\text{base}} = q$ and $\mathcal{I}_{\text{extra}} = (q, s_{\text{refine}})$, where $s_{\text{refine}}$ is the generated refinement. The information gain is:
\begin{equation}
\Delta(s_{\text{refine}}) = \text{LogitP}(a \mid q, s_{\text{refine}}) - \text{LogitP}(a \mid q)
\end{equation}
where $\Delta(s_{\text{refine}}) > 0$ indicates effective extraction of diagnostic-relevant information, and $\Delta(s_{\text{refine}}) \leq 0$ indicates failure to provide incremental knowledge.

Unlike C-MIG-D, where verbose external documents yield continuously varying magnitudes suited to truncated tanh mapping, refinement is model-generated and concise, making absolute magnitude less informative than relative standing within a batch. We therefore convert $\Delta(s_{\text{refine}})$ into rewards using a batch-rank mapping. Within each batch, we compute the median $\tilde{\Delta}^+$ over all positive-gain samples $\mathcal{P} = \{i : \Delta(s_{\text{refine}})^{(i)} > 0\}$ and reward only the upper half:
\begin{equation}
R_{\text{refine}}^{(i)} = \begin{cases} w_r, & \text{if } \Delta(s_{\text{refine}})^{(i)} \geq \tilde{\Delta}^+ > 0 \\ 0, & \text{otherwise} \end{cases}
\end{equation}
where $\tilde{\Delta}^+ = \text{median}(\{\Delta(s_{\text{refine}})^{(i)}\}_{i \in \mathcal{P}})$. This design dynamically adjusts the threshold according to batch difficulty, thereby avoiding over-incentivizing low-quality refinement.
% and eliminates hyperparameter sensitivity of continuous mappings\todo{what mean?}.
% An alternative asymmetric tanh variant is detailed in Appendix~\ref{app:refine-p-variants}.\todo{details}

C-MIG-D and C-MIG-R synergistically guide ``what to retrieve'' and ``how to refine,'' jointly aligning the model toward the diagnostic objective.

\subsection{Outcome Rewards and Training Objective}
\label{sec:training}
We introduce two outcome rewards: format reward and diagnosis reward. 
\paragraph{Format Reward}
$R_{\text{format}}$ checks that all tags are correct and that the number of \texttt{<evidence>} tags does not exceed the number of \texttt{<search>} tags, which prevents fabricated evidence, yielding $w_f$ if the format is correct and $0$ otherwise.
% $R_{\text{format}}$ verifies structural correctness: all tags properly closed and \texttt{<evidence>} count not exceeding \texttt{<search>} count. Pass yields $R_{\text{format}} = w_f$; fail yields $0$; severe violations incur $-\eta$.
\paragraph{Diagnostic Reward.}
$R_{\text{diag}}$ applies exact matching on normalized text within \texttt{<diagnosis>} tags: match yields $1$, otherwise $0$.
% \paragraph{Total Reward.}

The overall reward is as follows:
\begin{equation}
R_{\text{total}} = R_{\text{format}} + R_{\text{diag}} + R_{\text{doc}} + R_{\text{refine}}
\end{equation}
\paragraph{Policy Optimization.}
We adopt Group Relative Policy Optimization (GRPO)~\citep{guo2025deepseek}. For each sample $(q, a)$, we sample $G$ trajectories $\{o_i\}_{i=1}^G$. Let $r_t(\theta) = \frac{\pi_\theta(o_{i,t} \mid q, o_{i,<t})}{\pi_{\theta_{\text{old}}}(o_{i,t} \mid q, o_{i,<t})}$ denote the importance ratio. The objective is:
\begin{equation}
\begin{split}
\mathcal{J}(\theta) = \mathbb{E}\bigg[ \frac{1}{G} \sum_{i=1}^{G} \frac{1}{|o_i|} \sum_{t=1}^{|o_i|} \min\!\big( r_t \hat{A}_{i,t},\; \\
\text{clip}(r_t, 1{-}\epsilon, 1{+}\epsilon)\, \hat{A}_{i,t} \big) - \beta\, D_{\text{KL}}[\pi_\theta \| \pi_{\theta_0}] \bigg]
\end{split}
\label{eq:grpo}
\end{equation}
Here $\pi_{\theta_0}$ is the initial policy used only for KL regularization, distinct from the information-gain scorer $\pi_{\text{ref}}$. Advantages are estimated via group z-score normalization:
\begin{equation}
\hat{A}_i = \frac{R_i - \text{mean}(\{R_j\}_{j=1}^G)}{\text{std}(\{R_j\}_{j=1}^G) + \varepsilon}
\end{equation}
The same trajectory-level advantage is assigned to every token, i.e., $\hat{A}_{i,t} = \hat{A}_i$. Tokens within \texttt{<evidence>} tags are masked during loss computation, as they are environment-generated and should not receive policy gradients.

\begin{table*}[t]
\centering
\small 
\resizebox{\textwidth}{!}{%
\setlength{\aboverulesep}{0pt}%
\renewcommand{\arraystretch}{1.2}%
\begin{tabular}{lccccccccccccc}
\toprule
\multirow{2}{*}{Methods}
& \multicolumn{3}{c}{MedDDx-Plus (In)}
& \multicolumn{3}{c}{MedQA (OOD)}
& \multicolumn{3}{c}{MedXpertQA (OOD)}
& \multicolumn{3}{c}{RJUA (OOD)}
& \multirow{2}{*}{Overall Avg} \\
\cmidrule(lr){2-4} \cmidrule(lr){5-7} \cmidrule(lr){8-10} \cmidrule(lr){11-13}
& EM & KG & Avg & EM & KG & Avg & EM & KG & Avg & EM & KG & Avg & \\
\midrule
\multicolumn{14}{l}{\textit{Large Language Models (backbone-independent)}} \\
HuatuoGPT2-13B & 1.10 & 2.53 & 1.81 & 0.50 & 1.80 & 1.15 & 1.19 & 4.28 & 2.73 & 1.90 & 3.41 & 2.65 & 2.09 \\
Baichuan2-13B-Chat & 3.69 & 11.41 & 7.55 & 2.50 & 6.60 & 4.55 & 1.19 & 3.01 & 2.10 & 12.32 & 22.46 & 17.39 & 7.90 \\
Llama3-Med42-70B & 8.87 & 20.40 & 14.64 & 7.50 & 19.90 & 13.70 & 3.37 & 7.56 & 5.47 & 9.00 & 13.36 & 11.18 & 11.25 \\
Baichuan-M3-235B & 9.77 & 16.87 & 13.32 & 17.50 & 22.30 & 19.90 & 7.72 & 5.94 & 6.83 & 9.00 & 16.78 & 12.89 & 13.24 \\
DeepSeek-V4-Pro & 21.93 & 43.11 & 32.52 & 20.00 & 29.20 & 24.60 & 22.97 & 24.63 & 23.80 & 9.48 & 24.55 & 17.01 & 24.48 \\
GPT-5.2 & 20.04 & 42.85 & 31.45 & 12.00 & 25.20 & 18.60 & 8.12 & 13.31 & 10.71 & 8.53 & 31.94 & 20.24 & 20.25 \\
\midrule
\multicolumn{14}{l}{\textit{RAG-RL methods --- Qwen2.5-3B backbone}} \\
Qwen2.5-3B (Base) & 0.30 & 0.30 & 0.30 & 0.50 & 1.00 & 0.75 & 0.79 & 2.57 & 1.68 & 0.00 & 1.23 & 0.61 & 0.84 \\
Search-R1 & 54.34 & 57.61 & 55.98 & 3.50 & 11.90 & 7.70 & 0.99 & 1.82 & 1.41 & 7.11 & 19.53 & 13.32 & 19.60 \\
IGPO & 51.84 & 54.58 & 53.21 & \textbf{7.00} & 13.10 & 10.05 & 5.54 & 11.01 & 8.28 & 8.53 & 25.97 & 17.25 & 22.20 \\
AutoRefine & 48.65 & 51.47 & 50.06 & \textbf{7.00} & 12.00 & 9.50 & 3.96 & 8.51 & 6.23 & 8.53 & 24.35 & 16.44 & 20.56 \\
~~-~AutoRefine-HardSearch & 48.75 & 52.50 & 50.62 & 5.50 & 13.80 & 9.65 & 6.73 & \textbf{12.12} & 9.43 & \textbf{10.43} & 24.45 & 17.44 & 21.79 \\
~~-~AutoRefine-HardDoc & 52.94 & 58.52 & 55.73 & 5.50 & 11.40 & 8.45 & 4.36 & 7.52 & 5.94 & 7.11 & 18.77 & 12.94 & 20.77 \\
\rowcolor{cmigblue} \cmigstrut\textbf{C-MIG (Ours)} & \textbf{57.43} & \textbf{62.41} & \textbf{59.92} & 6.50 & \textbf{15.20} & \textbf{10.85} & \textbf{7.33} & 12.08 & \textbf{9.71} & 9.95 & \textbf{26.82} & \textbf{18.38} & \textbf{24.72} \\
\midrule
\multicolumn{14}{l}{\textit{RAG-RL methods --- Qwen2.5-7B backbone}} \\
Qwen2.5-7B (Base) & 2.99 & 5.80 & 4.39 & 0.00 & 2.90 & 1.45 & 0.59 & 2.65 & 1.62 & 0.00 & 0.76 & 0.38 & 1.96 \\
Search-R1 & 55.43 & 58.82 & 57.12 & 11.00 & 20.80 & 15.90 & 3.76 & 5.98 & 4.87 & 10.43 & 20.47 & 15.45 & 23.33 \\
IGPO & 52.84 & 56.21 & 54.53 & \textbf{13.00} & 22.60 & \textbf{17.80} & 3.76 & 7.22 & 5.49 & 8.53 & \textbf{28.63} & 18.58 & 24.10 \\
AutoRefine & 53.94 & 56.49 & 55.22 & 12.00 & 19.80 & 15.90 & 4.55 & 7.41 & 5.98 & 10.43 & 25.21 & 17.82 & 23.73 \\
~~-~AutoRefine-HardSearch & 58.62 & 63.51 & 61.06 & 10.50 & \textbf{23.40} & 16.95 & 5.15 & 11.17 & 8.16 & 9.48 & 23.51 & 16.50 & 25.66 \\
~~-~AutoRefine-HardDoc & 57.03 & 60.82 & 58.92 & 10.50 & 20.20 & 15.35 & 7.13 & 12.95 & 10.04 & 10.90 & 25.88 & 18.39 & 25.67 \\
\rowcolor{cmigblue} \cmigstrut\textbf{C-MIG (Ours)} & \textbf{60.82} & \textbf{65.12} & \textbf{62.97} & 11.00 & 20.40 & 15.70 & \textbf{7.52} & \textbf{13.39} & \textbf{10.46} & \textbf{11.37} & 27.49 & \textbf{19.43} & \textbf{27.14} \\
\bottomrule
\end{tabular}%
}
\caption{Clinical diagnostic performance on in-domain MedDDx-Plus and three out-of-domain datasets. The top block lists general-purpose and medical LLMs, whose results are backbone-independent; the lower blocks report RAG-RL methods on Qwen2.5-3B and Qwen2.5-7B. EM: exact match; KG: ICD-10 tree score; Avg: their mean. Bold marks the best value within each backbone block; C-MIG rows are shaded.}
\label{tab:main_result}
\end{table*}

\section{Experiments and Results}

\subsection{Experimental Settings}

% \subsubsection{Model and Baselines}
\textbf{Implementation Details.} We build C-MIG on the verl framework~\citep{sheng2025hybridflow} and experiment with three backbones: Qwen2.5-3B, Qwen2.5-7B~\citep{qwen2025qwen25technicalreport}, and Qwen3-4B~\citep{yang2025qwen3}. We train with the Adam optimizer at a learning rate of $6\times10^{-7}$ and a batch size of $64$ for $180$ steps on 4 H800 GPUs, allowing up to $T_{\max}=3$ retrieval turns per query. We train on MedDDx-Plus~\citep{fansi2022ddxplus}, which pairs patient symptom descriptions with standard diagnostic labels across multiple specialties; other training details are described in Appendix~\ref{appendix_setting}.

\textbf{Baselines.} We compare against these baselines: the Base Model, which performs direct inference without training; Search-R1~\citep{jin2025search}, which uses only an exact-match (EM) outcome reward for multi-round retrieval and reasoning; AutoRefine~\citep{shi2026search}, which extends Search-R1 with a refinement step rewarded by a hard-matching signal; and IGPO~\citep{wang2026information}, which uses the policy model itself to compute the turn-level information gain of retrieved documents. To tailor process rewards to clinical retrieval, we further build two hard process reward variants on top of AutoRefine, AutoRefine-HardSearch and AutoRefine-HardDoc, which give a hard-matching reward based on whether the generated queries or retrieved documents contain the ground-truth diagnosis. For reference, we also report four medical LLMs, HuatuoGPT2-13B~\citep{chen2023huatuogpt}, Baichuan2-13B-Chat~\citep{yang2023baichuan}, Llama3-Med42-70B~\citep{christophe2024med42}, and Baichuan-M3-235B~\citep{dou2026baichuan}, together with two general-purpose LLMs, DeepSeek-V4-Pro~\citep{xu2026deepseek} and GPT-5.2\footnote{https://developers.openai.com/api/docs/models/gpt-5.2}. We omit teacher-guided baselines, as they incur additional training overhead and a medical teacher can hardly instruct the policy to retrieve on its own, making them not directly comparable in our setting. 

\textbf{Benchmarks.} We evaluate on one in-domain benchmark, MedDDx-Plus, and three out-of-domain benchmarks: MedQA~\citep{jin2021disease}, which contains medical licensing exam questions; MedXpertQA~\citep{zuo2025medxpertqa}, which poses expert-level, interdisciplinary diagnostic questions; and RJUA~\citep{lyu2023rjua}, which collects real-world urology consultation dialogues. Dataset processing details and the evaluation protocol are provided in Appendix~\ref{appendix_preprocessing} and Appendix~\ref{appendix_metrics}.

\subsection{Main Results}

\begin{figure*}[t]
    \centering
    \includegraphics[width=\textwidth]{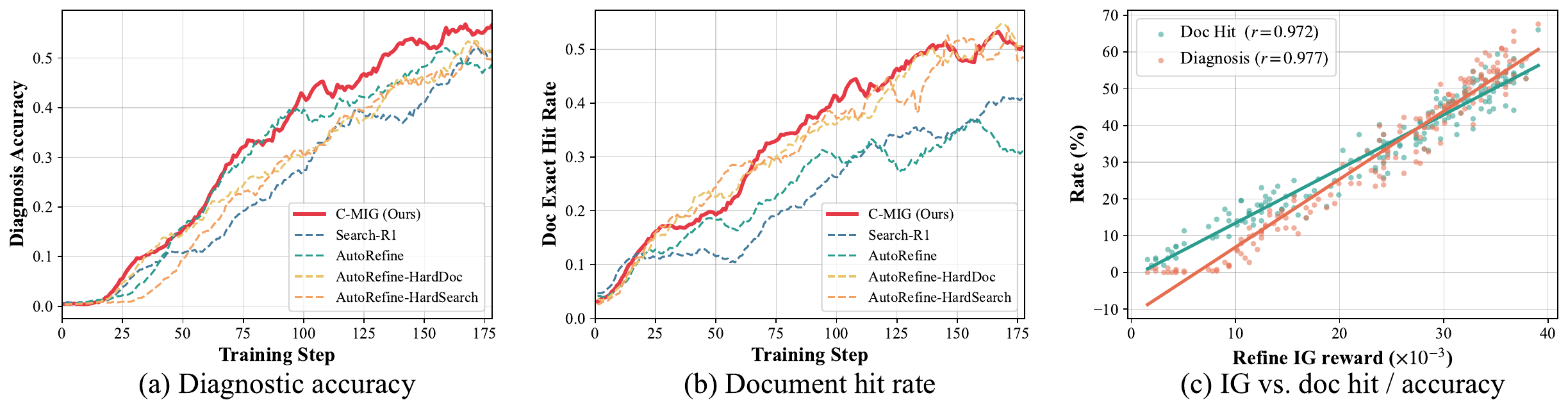}
    \caption{Training dynamics on Qwen2.5-3B: (a) diagnostic accuracy and (b) document hit rate over training; and (c) per-step document hit rate and diagnostic accuracy against the C-MIG refine-level information gain reward, showing that higher IG corresponds to higher Doc Hit and diagnostic accuracy (Pearson $r=0.972$ and $r=0.977$).}
    \label{fig:training_dynamics}
\end{figure*}

\subsubsection{Overall Performance}

Table~\ref{tab:main_result} compares C-MIG against baselines on the Qwen2.5-3B and Qwen2.5-7B backbones, with Qwen3-4B results reported in Appendix~\ref{appendix_main_result}. Under the same backbone and training protocol, C-MIG achieves the best overall average at both scales, reaching 24.72 on Qwen2.5-3B and 27.14 on Qwen2.5-7B. Against the strongest prior method IGPO, C-MIG improves on every dataset of the 3B backbone, MedDDx-Plus (+6.71), MedQA (+0.80), MedXpertQA (+1.43), and RJUA (+1.13), raising the in-domain score by 6.71 and the out-of-domain average by 1.12. The same comparison on 7B raises the in-domain score by 8.44 and the out-of-domain average by 1.24. Moreover, the baseline closest to C-MIG is AutoRefine-HardDoc, our own extension of AutoRefine; on the 7B backbone C-MIG still improves the in-domain score by 4.05 and the out-of-domain average by 0.60 over it. The gain is markedly larger in-domain than out-of-domain, as expected in this knowledge-intensive setting where the out-of-domain headroom is bounded less by the reward design than by whether the corpus covers the target disease. The out-of-domain gains therefore come mainly from stronger retrieval and better refinement that internalizes and extracts the useful evidence. Overall, the consistent margins across all four benchmarks and both scales further show that C-MIG is more robust than the baselines.

We further compare with the backbone-independent LLMs in Table~\ref{tab:main_result}. The powerful general-purpose DeepSeek-V4-Pro reaches 24.48 overall and GPT-5.2 20.25, while large medical models such as Baichuan-M3-235B (13.24) and Llama3-Med42-70B (11.25) trail behind. Despite having far fewer parameters, C-MIG already matches DeepSeek-V4-Pro at the 3B scale (24.72) and surpasses all of them at 7B (27.14). This shows that even top-tier models lack the specialized knowledge these benchmarks demand, and it demonstrates the value of high-quality RAG-RL, which grounds a small policy in retrievable medical evidence instead of parametric memorization.

\subsubsection{Consistency Analysis of Retrieval Quality and Diagnostic Accuracy}

To explain where the gains come from, we analyze the training process along the two dimensions C-MIG supervises, \emph{what to retrieve} and \emph{how to refine}.

\textbf{What to Retrieve.} As shown in Figure~\ref{fig:training_dynamics}(a) and (b), both the diagnostic accuracy and the document hit rate of C-MIG rise steadily throughout training, with its retrieval surpassing Search-R1, IGPO, and AutoRefine. AutoRefine instead plateaus or even fluctuates downward in the later stages, showing that a hard-matching process reward leaves little room for improvement in clinical diagnosis. To verify that better retrieval indeed drives better diagnosis, we compute the Pearson correlation between the two curves. As reported in Table~\ref{tab:pearson}, C-MIG reaches $r = 0.994$, substantially higher than IGPO at $0.963$ and AutoRefine-HardSearch at $0.966$, so each gain in retrieval quality translates into a synchronous gain in diagnostic accuracy. Notably, AutoRefine-HardDoc attains the highest document hit rate of $0.515$ yet only $0.518$ diagnostic accuracy, far below the $0.558$ of C-MIG. This indicates that supervising retrieval with a binary signal encourages shortcuts that maximize ground-truth inclusion without building robust reasoning, whereas the likelihood-based signal of C-MIG improves retrieval while preserving exploration.

\begin{table}[t]
\centering
\small
\begin{tabular}{lccc}
\toprule
Methods & Pearson $r$ & DA & Doc Hit \\
\midrule
Search-R1 & 0.988 & 0.513 & 0.403 \\
AutoRefine & 0.971 & 0.480 & 0.312 \\
IGPO & 0.963 & 0.522 & 0.375 \\
AutoRefine-HardSearch & 0.966 & 0.519 & 0.511 \\
AutoRefine-HardDoc & 0.991 & 0.518 & \textbf{0.515} \\
\midrule
\textbf{C-MIG} & \textbf{0.994} & \textbf{0.558} & 0.508 \\
\bottomrule
\end{tabular}
\caption{Pearson correlation ($r$) between diagnostic accuracy (DA) and document hit rate (Doc Hit) during training, alongside their final values.}
\label{tab:pearson}
\end{table}

\textbf{How to Refine.}
The analysis above aligns retrieval behavior with the diagnostic objective but does not yet test whether the refine-level information gain (IG) score itself corresponds to diagnostically relevant evidence. As shown in Figure~\ref{fig:training_dynamics}(c), we plot the per-step document hit rate and diagnostic accuracy against the refine-level IG reward across the 180 training steps, and both correlate strongly with IG, with $r=0.972$ for IG versus Doc Hit and $r=0.977$ for IG versus diagnosis. Binning steps into IG terciles, high-IG steps reach a $3.4\times$ higher document hit rate, $0.501$ versus $0.146$, and a $6.5\times$ higher diagnostic accuracy, $0.534$ versus $0.082$, than low-IG steps. Crucially, Doc Hit is computed by exact string matching against the gold diagnosis and is independent of the reference model, confirming that a high refinement information gain reflects more relevant evidence rather than an artifact of the scorer's internal preference. Appendices~\ref{appendix:training_stability} and~\ref{appendix_case_study} further provide a training stability analysis and case studies, including training cases where hard matching gives zero reward yet C-MIG credits etiologically relevant evidence.

\subsubsection{Effectiveness Analysis of the Frozen Reference Model}

In this section, we analyze why a frozen reference model is the right scorer for information gain in clinical diagnosis. Fixing the policy to Qwen2.5-3B, we vary the scorer from the policy itself to frozen references spanning 3B to 32B, including the general-purpose Baichuan2-13B and the medically specialized Baichuan-M2-32B~\citep{dou2025baichuan}, and report the per-dataset EM/KG average in Table~\ref{tab:ablation_posterior}, with full results in Appendix~\ref{appendix_ref_policy}.

\textbf{Stable Process Reward.} Every frozen reference beats policy self-scoring, with the 3B reference reaching 24.72 overall against 21.26. A self-scoring policy assigns shifting probabilities to identical documents as training proceeds, which is especially harmful in clinical diagnosis where diagnostic terms heavily overlap and the same evidence must retain a consistent value. A frozen reference keeps the criterion time-invariant and makes information gain comparable across steps.

\textbf{No Medical Teacher Needed.} Unlike teacher-guiding methods, whose reliability is bounded by how much medicine the teacher knows, C-MIG recasts process-reward assignment into a basic likelihood comparison that any base model can carry out. Scaling Qwen2.5 from 3B to 32B moves the overall score only from 24.72 to 25.17, and the medically specialized Baichuan-M2-32B reaches 25.16, no better than the general-purpose Baichuan2-13B at 25.20. Judging whether evidence raises the likelihood of a diagnosis is therefore capability-agnostic rather than a task demanding medical expertise.

\textbf{Low Training Overhead.} C-MIG needs only one extra forward pass through a small Qwen2.5-3B reference, kept in GPU memory by a persistent worker pool and fed through dynamic token-budget batching. It adds only 3.4\% to step time and 6.8 GB per GPU, under 5\% of the total training wall-clock. The reference is never loaded at inference, so deployment matches a single-model RAG. Full measurements are in Appendix~\ref{appendix_compute_overhead}.

\begin{table}[t]
\centering
\small
\resizebox{0.5\textwidth}{!}{%
\begin{tabular}{lccccc}
\toprule
Scorer Model & MD & MQA & MX & RJUA & Overall \\
\midrule
Policy (self) & 46.36 & 8.85 & 8.23 & 21.61 & 21.26 \\
\hdashline\noalign{\vskip \tblsep}
Ref: Qwen2.5-3B & 59.92 & 10.85 & 9.71 & 18.38 & 24.72 \\
Ref: Qwen2.5-7B & 61.02 & 9.80 & 10.54 & 18.15 & 24.88 \\
Ref: Qwen2.5-32B & 61.67 & 9.47 & 10.37 & 19.17 & 25.17 \\
Ref: Baichuan2-13B & 63.54 & 10.15 & 8.12 & 18.99 & \textbf{25.20} \\
Ref: Baichuan-M2-32B & 60.96 & 8.80 & 7.93 & 22.95 & 25.16 \\
\bottomrule
\end{tabular}%
}
\caption{Scorer selection for information gain, policy fixed to Qwen2.5-3B. MD: MedDDx-Plus; MQA: MedQA; MX: MedXpertQA.}
\label{tab:ablation_posterior}
\end{table}

\begin{table}[t]
\centering
\small
\resizebox{0.5\textwidth}{!}{%
\begin{tabular}{lccccc}
\toprule
Methods & MD & MQA& MX& RJUA& Overall \\
\midrule
Base & 50.06 & 9.50 & 6.23 & 16.44 & 20.56 \\[\tblsep]
\hdashline\noalign{\vskip \tblsep}
+ Doc & 57.55 & 10.10 & 9.32 & 17.39 & 23.59 \\
+ Refine & 56.75 & 9.05 & 7.03 & \textbf{19.43} & 23.07 \\[\tblsep]
\hline\noalign{\vskip \tblsep}
\textbf{C-MIG} & \textbf{59.92} & \textbf{10.85} & \textbf{9.71} & 18.38 & \textbf{24.72} \\
\bottomrule
\end{tabular}%
}
\caption{Ablation study on the AutoRefine base model, where +Doc and +Refine add information gain from retrieved documents and refined content, respectively.}
\label{tab:ablation}
\end{table}

% \begin{table}[t]
% \centering
% \small
% % \resizebox{\textwidth}{!}{%
% \begin{tabular}{lccccc}
% \toprule
% Granularity & MD & MQA& MX & RJUA& Overall \\
% \midrule
% IGPO (Turn) & 54.53 & \textbf{17.80} & 5.74 & 18.58 & 24.16 \\
% C-MIG-D (Seq) & \textbf{55.73} & 16.45 & \textbf{9.72} & \textbf{21.18} & \textbf{25.77} \\
% \bottomrule
% \end{tabular}%
% % }
% \caption{Comparison of different reward granularity designs.}
% \label{tab:ablation_granularity_v2}
% \end{table}

\subsection{Ablation Study}

To validate the effectiveness of each component, we adopt Qwen2.5-3B as the base model and progressively incorporate modules on top of AutoRefine: (1) +Doc, adding information gain from retrieved documents; (2) +Refine, adding information gain from refined content. The full C‑MIG integrates both components. Results are presented in Table \ref{tab:ablation}, with full EM/KG results on Qwen2.5-3B and Qwen2.5-7B in Appendix \ref{appendix_ablation_study}. 

The addition of each component improves overall performance, validating the effectiveness of our multi‑view information gain mechanism. The gains from multi‑view information gain are particularly substantial: +Doc and +Refine yield improvements of 3.03 and 2.51, respectively. Computing information gain from retrieved documents enhances retrieval quality, whereas computing it from refined content improves knowledge summarization capability. Together, they supervise two orthogonal capability dimensions: ``retrieval planning'' and ``knowledge digestion.'' We further analyze rollout-level versus turn-level reward granularity in Appendix~\ref{appendix_reward_level} and the scalability of multi-view information gain across backbones in Appendix~\ref{appendix_scalability}.

\section{Conclusion}

We propose \textbf{C-MIG}, a multi-view information gain-based RAG-RL framework for clinical diagnosis. C-MIG rewards two complementary steps, \emph{what to retrieve} and \emph{how to refine}, by the gain each brings to the predicted likelihood of the ground-truth answer under a base, frozen reference model, yielding a continuous and reliable process reward without hard matching or a strong teacher. Experiments on four medical benchmarks confirm its effectiveness at low training cost and no inference-time overhead. We hope C-MIG inspires future work on reliable process rewards for retrieval-augmented clinical reasoning.

\section*{Limitations}

This work is subject to several limitations. Our method relies on a frozen reference model, the evaluation is restricted to text-only medical question answering and diagnostic dialogues without incorporating multi-modal evidence such as medical images, and the benchmarks used may not fully reflect the diversity of patient populations across different regions, languages, and healthcare systems. Accordingly, the experimental results presented in this paper should be interpreted as methodological evidence for the effectiveness of retrieval-augmented diagnostic reasoning, rather than as evidence of readiness for clinical deployment.
Two further caveats concern the data and the reward signal. Our out-of-domain benchmarks are converted into open-ended English question answering with a strong LLM, and although the rewriting only replaces multiple-choice phrasing and the filtering ratio is uniform across task types, residual conversion bias cannot be fully excluded. In addition, a process reward derived from a frozen reference model can reproduce that model's clinical blind spots, and we report benchmark-level diagnostic accuracy and retrieval relevance rather than expert-adjudicated clinical quality.
All public models and baselines used in this work are employed under their original licenses or service terms, while proprietary large language models are accessed through their official services. Future work will focus on scaling C-MIG to larger and multi-modal foundation models, conducting broader cross-institutional and cross-lingual validation, and incorporating evaluations by professional clinicians together with safer evaluation protocols.

% This document does not cover the content requirements for ACL or any
% other specific venue.  Check the author instructions for
% information on
% maximum page lengths, the required ``Limitations'' section,
% and so on.

% \section*{Acknowledgments}

% This document has been adapted
% by Steven Bethard, Ryan Cotterell and Rui Yan
% from the instructions for earlier ACL and NAACL proceedings, including those for
% ACL 2019 by Douwe Kiela and Ivan Vuli\'{c},
% NAACL 2019 by Stephanie Lukin and Alla Roskovskaya,
% ACL 2018 by Shay Cohen, Kevin Gimpel, and Wei Lu,
% NAACL 2018 by Margaret Mitchell and Stephanie Lukin,
% Bib\TeX{} suggestions for (NA)ACL 2017/2018 from Jason Eisner,
% ACL 2017 by Dan Gildea and Min-Yen Kan,
% NAACL 2017 by Margaret Mitchell,
% ACL 2012 by Maggie Li and Michael White,
% ACL 2010 by Jing-Shin Chang and Philipp Koehn,
% ACL 2008 by Johanna D. Moore, Simone Teufel, James Allan, and Sadaoki Furui,
% ACL 2005 by Hwee Tou Ng and Kemal Oflazer,
% ACL 2002 by Eugene Charniak and Dekang Lin,
% and earlier ACL and EACL formats written by several people, including
% John Chen, Henry S. Thompson and Donald Walker.
% Additional elements were taken from the formatting instructions of the \emph{International Joint Conference on Artificial Intelligence} and the \emph{Conference on Computer Vision and Pattern Recognition}.


\bibliography{custom}

\clearpage
\appendix

\twocolumn[%
\begin{center}
\small
\resizebox{\textwidth}{!}{%
\setlength{\aboverulesep}{0pt}%
\renewcommand{\arraystretch}{1.2}%
\begin{tabular}{lccccccccccccc}
\toprule
\multirow{2}{*}{Methods}
& \multicolumn{3}{c}{MedDDx-Plus (In)}
& \multicolumn{3}{c}{MedQA (OOD)}
& \multicolumn{3}{c}{MedXpertQA (OOD)}
& \multicolumn{3}{c}{RJUA (OOD)}
& \multirow{2}{*}{Overall Avg} \\
\cmidrule(lr){2-4} \cmidrule(lr){5-7} \cmidrule(lr){8-10} \cmidrule(lr){11-13}
& EM & KG & Avg & EM & KG & Avg & EM & KG & Avg & EM & KG & Avg & \\
\midrule
\multicolumn{14}{l}{\textit{Large Language Models (backbone-independent)}} \\
HuatuoGPT2-13B & 1.10 & 2.53 & 1.81 & 0.50 & 1.80 & 1.15 & 1.19 & 4.28 & 2.73 & 1.90 & 3.41 & 2.65 & 2.09 \\
Baichuan2-13B-Chat & 3.69 & 11.41 & 7.55 & 2.50 & 6.60 & 4.55 & 1.19 & 3.01 & 2.10 & 12.32 & 22.46 & 17.39 & 7.90 \\
Llama3-Med42-70B & 8.87 & 20.40 & 14.64 & 7.50 & 19.90 & 13.70 & 3.37 & 7.56 & 5.47 & 9.00 & 13.36 & 11.18 & 11.25 \\
Baichuan-M3-235B & 9.77 & 16.87 & 13.32 & 17.50 & 22.30 & 19.90 & 7.72 & 5.94 & 6.83 & 9.00 & 16.78 & 12.89 & 13.24 \\
DeepSeek-V4-Pro & 21.93 & 43.11 & 32.52 & 20.00 & 29.20 & 24.60 & 22.97 & 24.63 & 23.80 & 9.48 & 24.55 & 17.01 & 24.48 \\
GPT-5.2 & 20.04 & 42.85 & 31.45 & 12.00 & 25.20 & 18.60 & 8.12 & 13.31 & 10.71 & 8.53 & 31.94 & 20.24 & 20.25 \\
\midrule
\multicolumn{14}{l}{\textit{RAG-RL methods --- Qwen3-4B backbone}} \\
Qwen3-4B (Base) & 0.00 & 7.76 & 3.88 & 0.00 & 6.60 & 3.30 & 0.20 & 1.54 & 0.87 & 0.00 & 3.51 & 1.76 & 2.45 \\
Search-R1 & \textbf{53.54} & \textbf{59.76} & \textbf{56.65} & 8.50 & 16.70 & 12.60 & 6.34 & 10.02 & 8.18 & 7.11 & 21.61 & 14.36 & 22.95 \\
IGPO & 48.35 & 57.27 & 52.81 & 8.00 & 15.80 & 11.90 & 2.18 & 4.67 & 3.42 & 12.80 & 22.37 & 17.59 & 21.43 \\
AutoRefine & 45.76 & 57.37 & 51.56 & 8.00 & 16.60 & 12.30 & 4.75 & 7.88 & 6.31 & \textbf{15.17} & 24.64 & \textbf{19.91} & 22.52 \\
~~-~AutoRefine-HardSearch & 48.55 & 58.21 & 53.38 & \textbf{10.00} & \textbf{18.60} & \textbf{14.30} & 3.37 & 5.50 & 4.44 & 10.43 & 22.37 & 16.40 & 22.13 \\
~~-~AutoRefine-HardDoc & 49.45 & 57.61 & 53.53 & 8.00 & 15.80 & 11.90 & 1.58 & 4.00 & 2.79 & 9.95 & 19.15 & 14.55 & 20.69 \\
\rowcolor{cmigblue} \cmigstrut\textbf{C-MIG (Ours)} & 50.75 & 58.72 & 54.74 & 9.00 & 17.20 & 13.10 & \textbf{7.92} & \textbf{12.87} & \textbf{10.40} & 9.48 & \textbf{28.15} & 18.81 & \textbf{24.26} \\
\bottomrule
\end{tabular}%
}
\captionof{table}{Main results on the Qwen3-4B backbone. The top block repeats the backbone-independent LLM references; the lower block reports RAG-RL methods trained on Qwen3-4B. EM: exact match; KG: ICD-10 tree score; Avg: their mean. Bold marks the best within the Qwen3-4B RAG-RL block; our C-MIG row is shaded.}
\label{tab:main_results_qwen3_4b}
\end{center}
\vspace{0.8em}
]

% \section{Refine-P Reward Mapping Variant}
% \label{app:refine-p-variants}

% This appendix presents an alternative Refine-P reward mapping based on asymmetric hyperbolic tangent functions. The half-absolute scheme in the main text achieves adaptive thresholds through batch-relative ranking but discards magnitude information: a high-quality refine with $\Delta = 5.0$ and a barely qualifying one with $\Delta = 0.1$ receive identical rewards. An alternative preserves magnitude and applies asymmetric penalties:
% \begin{equation}
% R_{\text{refine}}^{\text{asym}} = \begin{cases} w_r \cdot \tanh(\alpha^+ \cdot \Delta_{\text{refine}}), & \text{if } \Delta_{\text{refine}} \geq 0 \\ w_r \cdot \beta \cdot \tanh(\alpha^- \cdot \Delta_{\text{refine}}), & \text{if } \Delta_{\text{refine}} < 0 \end{cases}
% \end{equation}
% where $\alpha^+$ is the positive-region sensitivity (smaller, avoiding over-reward of weak gains), $\alpha^- > \alpha^+$ makes the negative region steeper, $\beta > 1$ amplifies penalties on misleading refines, and $w_r$ is the global weight. This embodies the clinical principle of ``primum non nocere'': harm from misleading summarization exceeds the benefit from useful information. The bounded $\tanh$ output prevents extreme values from destabilizing training while preserving richer gradient signals than binarization.

\section{Main Result on Qwen3-4B}
\label{appendix_main_result}

We further evaluate C-MIG on the \textbf{Qwen3-4B} backbone under our updated training stack, comparing against Search-R1, IGPO, and the AutoRefine family (base, HardSearch, HardDoc) under an identical retrieval corpus and evaluation protocol; results are reported in Table~\ref{tab:main_results_qwen3_4b}. C-MIG attains the best overall score ($24.26$), ahead of Search-R1 ($22.95$), AutoRefine ($22.52$), AutoRefine-HardSearch ($22.13$), IGPO ($21.43$), and AutoRefine-HardDoc ($20.69$), confirming that the gains observed on the Qwen2.5 backbones transfer to a stronger, more recent backbone.

The advantage is most pronounced on the hardest out-of-domain benchmark, MedXpertQA. C-MIG is the only method above $10$ there, reaching an average of $10.40$ and improving over the strong baseline Search-R1 at $8.18$ by $27\%$ relative, while the hard-matching variants and IGPO remain below $4.5$, with HardSearch at $4.44$, IGPO at $3.42$ and HardDoc at $2.79$. This reproduces our central observation: when retrieved evidence seldom contains the gold diagnosis verbatim, binary hard rewards degenerate to a near-zero signal, whereas the continuous information gain of C-MIG still credits semantically relevant evidence and yields a usable learning signal.

Beyond this benchmark, the baselines are uneven across datasets, each leading somewhere and falling back elsewhere, whereas C-MIG is strong on both sides of the domain split. It attains the best out-of-domain average at $14.10$, ahead of the next best at $12.84$, while staying within $1.91$ of the best in-domain score. Its overall lead therefore does not rest on fitting a single benchmark, indicating that multi-view information gain provides a more robust optimization target than single-view or hard-matching rewards.

\begin{table*}[t]
\centering
\small
\renewcommand{\arraystretch}{1.15}
\setlength{\tabcolsep}{4pt}
\resizebox{\textwidth}{!}{%
\begin{tabular}{l l ccc ccc ccc ccc c}
\toprule
\multirow{2}{*}{Backbone} & \multirow{2}{*}{Method}
& \multicolumn{3}{c}{MedDDx-Plus}
& \multicolumn{3}{c}{MedQA}
& \multicolumn{3}{c}{MedXpertQA}
& \multicolumn{3}{c}{RJUA}
& \multirow{2}{*}{Overall Avg} \\
\cmidrule(lr){3-5} \cmidrule(lr){6-8} \cmidrule(lr){9-11} \cmidrule(lr){12-14}
& & EM & KG & Avg & EM & KG & Avg & EM & KG & Avg & EM & KG & Avg & \\
\midrule
\multirow{4}{*}{Qwen2.5-3B}
& AutoRefineBase & 48.65 & 51.47 & 50.06 & 7.00 & 12.00 & 9.50  & 3.96 & 8.51 & 6.23 & 8.53  & 24.35 & 16.44 & 20.56 \\
& + Doc         & 56.43 & 58.66 & 57.55 & 7.00 & 13.20 & 10.10 & 6.93 & 11.72 & 9.32 & 9.00  & 25.78 & 17.39 & 23.59 \\
& + Refine      & 55.03 & 58.46 & 56.75 & 5.50 & 12.60 & 9.05  & 4.36 & 9.70  & 7.03 & \textbf{11.85} & \textbf{27.01} & \textbf{19.43} & 23.07 \\
& \textbf{C-MIG}   & \textbf{57.43} & \textbf{62.41} & \textbf{59.92} & 6.50 & \textbf{15.20} & \textbf{10.85} & \textbf{7.33} & \textbf{12.08} & \textbf{9.71} & 9.95 & 26.82 & 18.38 & \textbf{24.72} \\
\midrule
\multirow{4}{*}{Qwen2.5-7B}
& AutoRefineBase & 53.94 & 56.49 & 55.22 & 12.00 & 19.80 & 15.90 & 4.55 & 7.41  & 5.98 & 10.43 & 25.21 & 17.82 & 23.73 \\
& + Doc         & 53.74 & 57.73 & 55.73 & 11.50 & 21.40 & 16.45 & 6.93 & 12.51 & 9.72 & \textbf{14.69} & \textbf{27.68} & \textbf{21.18} & 25.77 \\
& + Refine      & 60.32 & 64.87 & 62.59 & \textbf{12.50} & 21.00 & 16.75 & 6.34 & 11.21 & 8.78 & 10.90 & 27.30 & 19.10 & 26.81 \\
& \textbf{C-MIG}    & \textbf{60.82} & \textbf{65.12} & \textbf{62.97} & 11.00 & 20.40 & 15.70 & \textbf{7.52} & \textbf{13.39} & \textbf{10.46} & 11.37 & 27.49 & 19.43 & \textbf{27.14} \\
\bottomrule
\end{tabular}
}
\caption{Ablation study of C-MIG components on in-domain (MedDDx-Plus) and out-of-domain (MedQA, MedXpertQA, RJUA) benchmarks. Both +Doc and +Refine improve over the AutoRefine base, and the full C-MIG performs best on both backbones.}
\label{tab:cmig-ablation}
\end{table*}

\section{Ablation Study}
\label{appendix_ablation_study}

We conduct the component ablation on both the Qwen2.5-3B (Base) and Qwen2.5-7B (Base) backbones, progressively adding the two information-gain views on top of AutoRefine. The full EM/KG results are shown in Table~\ref{tab:cmig-ablation}.

Each view contributes on its own and the two are complementary. On Qwen2.5-3B, +Doc and +Refine raise the overall average by 3.03 and 2.51 over the AutoRefine base (20.56), reaching 23.59 and 23.07 respectively, and combining them yields the best score of 24.72. The same pattern holds on Qwen2.5-7B, where +Doc and +Refine improve the base (23.73) by 2.04 and 3.08 to 25.77 and 26.81, and the full C-MIG reaches 27.14. The relative weight of the two views shifts with scale: retrieval supervision (+Doc) contributes more on the smaller 3B model, whereas refinement supervision (+Refine) contributes more on the larger 7B model, consistent with the observation that higher-capacity policies benefit more from distilled, semantically concise evidence. Across both backbones the full model surpasses every single-view variant, confirming that the two views supervise orthogonal capabilities, retrieval planning and knowledge digestion, and that their combination is essential to the effectiveness of C-MIG.

\section{Experiments Setting}
\label{appendix_setting}
\subsection{Hyperparameter}
% We set $w_d = X, \alpha_d = X, w_r = X, w_f = X$. The learning rate = X, 

Training is performed on the verl framework with the Adam optimizer~\citep{kingma2014adam}, using a learning rate of $6\times 10^{-7}$ and a batch size of $64$ for $180$ steps. We sample a group of $G = 4$ rollouts with a maximum response length of $2{,}048$ tokens. The clip ratio is set to $0.2$, the KL regularization coefficient $\beta$ to $0.001$ for training stability.

\paragraph{Reward Weights.}
We set the format reward weight $w_f = 1$, diagnostic reward weight to $1$ (exact match), and hard retrieval reward weight to $0.1$. For the multi-turn retrieval information gain, we set weight $w_d = 1$ with sensitivity parameter $\alpha_d = 1$, and the refine reward weight $w_r = 0.1$. The elevated weight for information gain ($w_d = 1$) is motivated by the observation that in medical retrieval scenarios, reward signals are inherently sparse and models tend to collapse into single-turn retrieval without a sufficient process reward.

\paragraph{Retrieval Configuration.}
For retrieval-augmented generation, we use the medical retrieval corpus from \citet{zheng2025end}, which aggregates approximately 23.9M PubMed abstracts, 3.31M biomedical Wikipedia entries, 18 authoritative medical textbooks, and around 1,419 healthcare web resources (e.g., NCBI, NHS, Mayo Clinic, MedlinePlus, MSD Manuals), providing comprehensive clinical knowledge coverage. We employ BM25~\citep{robertson1994some} as the retriever with a Top-$k$ of $3$ and allow up to $T_{\max} = 3$ retrieval turns per response. All experiments are conducted on 4 H800 GPUs.

\subsection{Train and Test Datasets}
We use MedDDx-Plus~~\citep{fansi2022ddxplus} as the training dataset, which covers clinical diagnostic scenarios across multiple specialties, with each sample consisting of a patient symptom description and its corresponding standard diagnostic label. We uniformly sample 8{,}998 instances for training and 1{,}003 instances for validation from the original dataset.

To comprehensively evaluate the generalization capability of the model, we conduct evaluations on one in-domain (ID) dataset and three out-of-domain (OOD) datasets. The in-domain evaluation is performed on \textbf{MedDDx-Plus}, whose test subset is drawn from the same source as the training set and contains 1{,}003 evaluation instances. Since the original OOD benchmarks are in multiple-choice or non-English formats, we employ GPT-5.2 to preprocess them into a unified open-ended English QA format suitable for diagnostic evaluation. The three OOD datasets are: (1) \textbf{MedQA}~\citep{jin2021disease}, originally a multiple-choice medical examination benchmark containing many pharmacology-related questions; we use GPT-5.2 to filter instances suitable for disease diagnosis and convert them from multiple-choice to open-ended QA format, yielding 200 instances; (2) \textbf{MedXpertQA}~\citep{zuo2025medxpertqa}, an expert-level multiple-choice medical dataset of 505 instances comprising challenging interdisciplinary diagnostic questions, converted to QA format via GPT-5.2; and (3) \textbf{RJUA}~~\citep{lyu2023rjua}, a real-world Chinese urology consultation dialogue dataset of 211 instances, characterized by highly colloquial patient descriptions and frequent multi-disease co-diagnosis, which we translate into professional English format using GPT-5.2.
% rendering it the most clinically realistic among the evaluation benchmarks. 
% The external knowledge base is constructed from clinical guideline resources\todo{ref}, with E5‑base‑v2\todo{ref} employed as the retrieval model.

\begin{table*}[t]
\centering
\small
\begin{tabular}{p{2.2cm} p{3.4cm} p{6.6cm} c}
\toprule
Dataset & Original format & Conversion to open-ended English QA & Eval $N$ \\
\midrule
MedDDx-Plus & Simulated doctor--patient dialogues (DDXPlus, 48 diseases) & None; the target is the standard diagnosis name & 1{,}003 \\
MedQA & 4-option USMLE-style MCQ & GPT prompt rewrite (MCQ$\rightarrow$open-ended) and retain only diagnosis-oriented items & 200 \\
MedXpertQA & 10-option expert MCQ (2{,}450 orig.) & GPT suitability filtering ($-500$) and prompt rewrite & 505 \\
RJUA & Real Chinese urology consultations & GPT zh$\rightarrow$en translation, extract the primary diagnosis, and drop expert context & 211 \\
\bottomrule
\end{tabular}
\caption{Out-of-domain preprocessing pipeline. Every dataset is evaluated with EM and KG; MedDDx-Plus is in-domain and used as-is. ``Eval $N$'' is the number of evaluation instances reported in the main results.}
\label{tab:preprocessing}
\end{table*}

\subsection{Dataset Preprocessing Details}
\label{appendix_preprocessing}
Since the original out-of-domain benchmarks are multiple-choice or non-English, we convert them into a unified open-ended English QA format with GPT-5.2. Table~\ref{tab:preprocessing} summarizes the pipeline; all datasets are evaluated with EM and KG.

\paragraph{MedQA type filtering.}
We use GPT-5.2 to classify each MedQA item into three categories. Diagnosis-oriented items ask for a disease or condition name, retrieval-suitable but non-diagnosis items ask for anatomical structures or physical signs, and unsuitable items concern pharmacology, biochemistry, or dosage calculations. We retain only the $787$ diagnosis-oriented items and discard the $1{,}743$ retrieval-suitable but non-diagnosis and $1{,}131$ unsuitable ones, because the final output of a RAG diagnostic consultation should be a disease name rather than a drug dose or a biochemical mechanism.

\paragraph{MedXpertQA quantity filtering.}
The original expert test set contains $2{,}450$ items, from which GPT-5.2 removes the $500$ whose answers are descriptive sentences rather than concise medical labels, leaving $1{,}950$. This $20.4\%$ removal ratio is nearly uniform across the three task types, with Basic Science dropping from $654$ to $521$, Diagnosis from $1{,}050$ to $836$, and Treatment from $746$ to $593$, so filtering does not introduce a task-type selection bias. The $505$-instance evaluation split we report on is kept without filtering.

\paragraph{MCQ-to-QA rewriting rules.}
The question rewriting performs only a minimal, meaning-preserving transformation:
\begin{itemize}
\item Only substitute multiple-choice phrasing with open-ended phrasing, e.g., ``Which of the following is the most likely diagnosis?'' $\rightarrow$ ``What is the most likely diagnosis?''.
\item Do not change any clinical detail, patient description, laboratory value, or sentence structure; expressions such as ``this patient'' are kept verbatim.
\item The rewritten question is open-ended, so the model can no longer rely on the option list to answer.
\end{itemize}

\paragraph{Exact rewriting prompt.}
For full reproducibility, the exact system prompt used by our MCQ-to-QA converter is shown below. Each question is additionally wrapped in a short per-item instruction: ``Rewrite this multiple-choice question into a natural open-ended question. Only change the question phrasing, keep all clinical details unchanged. Output ONLY the rewritten question.''

\begin{tcolorbox}[promptbox, title=System Prompt for MCQ-to-Open-ended Rewriting]
You are a medical question rewriter. Your task is to convert multiple-choice question phrasing to open-ended question phrasing.\\[2pt]
RULES:\\
1. ONLY do simple word substitution to convert the question to open-ended form.\\
\hspace*{1em}-- ``Which of the following is the most likely\ldots'' $\to$ ``What is the most likely\ldots''\\
\hspace*{1em}-- ``Which of the following is the most appropriate\ldots'' $\to$ ``What is the most appropriate\ldots''\\
\hspace*{1em}-- ``Which of the following is the best\ldots'' $\to$ ``What is the best\ldots''\\
\hspace*{1em}-- ``Which of the following would\ldots'' $\to$ ``What would\ldots''\\
\hspace*{1em}-- ``Which of the following can\ldots'' $\to$ ``What can\ldots''\\
2. Do NOT change any clinical details, patient descriptions, or sentence structure.\\
3. Keep ``this patient'' as-is (do NOT change to ``this boy'' etc.).\\
4. Keep all lab values, symptoms, history exactly as-is.\\
5. Output ONLY the rewritten question text, nothing else.
\end{tcolorbox}

\paragraph{Key design decisions.}
\begin{itemize}
\item Removing the options makes the task strictly harder, since the model can no longer eliminate answers, yielding a conservative evaluation.
\item Rewriting changes only the question form, never the clinical content.
\item MedQA type filtering aligns the task with disease-name diagnosis, matching the RAG consultation setting.
\item MedXpertQA filtering is proportionally uniform across task types, so no task family is over- or under-represented.
\item RJUA translation changes only the surface language and removes the expert context, so the model sees only the patient description.
\item \textbf{Fairness.} All methods are evaluated on exactly the same converted data, so any conversion artifact affects every method equally.
\end{itemize}

\subsection{Evaluation Metrics}
\label{appendix_metrics}

\paragraph{Outcome Metrics.}
In our setting each model outputs a single diagnosis per case. We adopt two diagnostic accuracy metrics with complementary granularity:

\begin{itemize}
\item \textbf{Exact Match (EM).} Returns 1 if the normalized prediction exactly matches the ground truth after lowercasing and punctuation removal. EM reflects the strictest diagnostic correctness criterion.

\item \textbf{Knowledge Graph Score (KG).} A continuous metric based on ICD-10 tree distance:
\begin{equation}
\text{KG} = \max\left(0,\; 1 - \alpha \cdot d(\hat{c}, c^*)\right), \quad \alpha = 0.2
\end{equation}
where $d(\hat{c}, c^*)$ is the shortest path length between the predicted and ground-truth ICD-10 codes through their Lowest Common Ancestor. Free-text diagnoses are mapped to ICD-10 codes via a cascaded strategy: regex extraction of explicit codes, exact dictionary lookup with inverted-index scoring over WHO ICD-10 descriptions, and an LLM fallback for unresolved cases; failed mappings receive a score of 0. KG rewards correct disease-family identification while remaining agnostic to within-category severity differences—a deliberate design choice complemented by strict EM evaluation.
\end{itemize}
\vspace{-6pt}
\noindent

We report EM and KG as the primary evaluation metrics across all methods, as both carry interpretable clinical semantics: EM ensures diagnostic precision, while KG captures clinically meaningful partial correctness through taxonomic proximity in the standardized ICD-10 coding system.

\paragraph{Process Reward Baselines.}
We implement hard process reward variants upon AutoRefine~~\citep{shi2026search}. Given the gold answer $a = (a_1, \ldots, a_L)$ of length $L$, all variants share a unified binary form:
\begin{equation}
R_{\text{process}} = \mathbb{I}\left[a \subseteq \mathcal{I}_t\right]
\end{equation}
where $\mathbb{I}$ denotes the binary indicator operator that takes the value of either $0$ or $1$ and $\mathcal{I}_t$ denotes the intermediate content at turn $t$. Depending on the variant, $\mathcal{I}_t$ corresponds to: the refined knowledge summary (AutoRefine base), the generated search query (AutoRefine-HardSearch), or the retrieved documents (AutoRefine-HardDoc). These binary signals suffer from the same sparsity as EM: they cannot distinguish near-miss from irrelevant results.

\paragraph{Soft Outcome Reward Baseline.}
We also extend AutoRefine~\citep{shi2026search} with a soft outcome reward that addresses the sparse reward problem in clinical diagnosis, where semantically equivalent disease names receive zero reward under exact matching. \textbf{AutoRefine-ICDTree} computes a hierarchical soft reward based on node distance in the ICD~\citep{maercker2013proposals} disease taxonomy.

\paragraph{Document Hit (Doc\_Hit).}
We use Doc\_Hit as the retrieval quality evaluation metric, defined identically to AutoRefine-HardDoc but aggregated over all turns:
\begin{equation}
\text{Doc\_Hit} = \mathbb{I}\left[a \subseteq \bigcup\nolimits_{t=1}^{T} \mathcal{D}_t\right]
\end{equation}
This metric measures whether the model successfully retrieves diagnostically relevant evidence across the entire reasoning process, serving as a necessary prerequisite for accurate diagnosis.

\section{Training Stability Analysis}
\label{appendix:training_stability}

We provide a comprehensive analysis of training stability metrics for all methods on Qwen2.5-3B, including late-stage volatility, gradient norm, and response length.

\begin{figure*}[t]
    \centering
    \includegraphics[width=\textwidth]{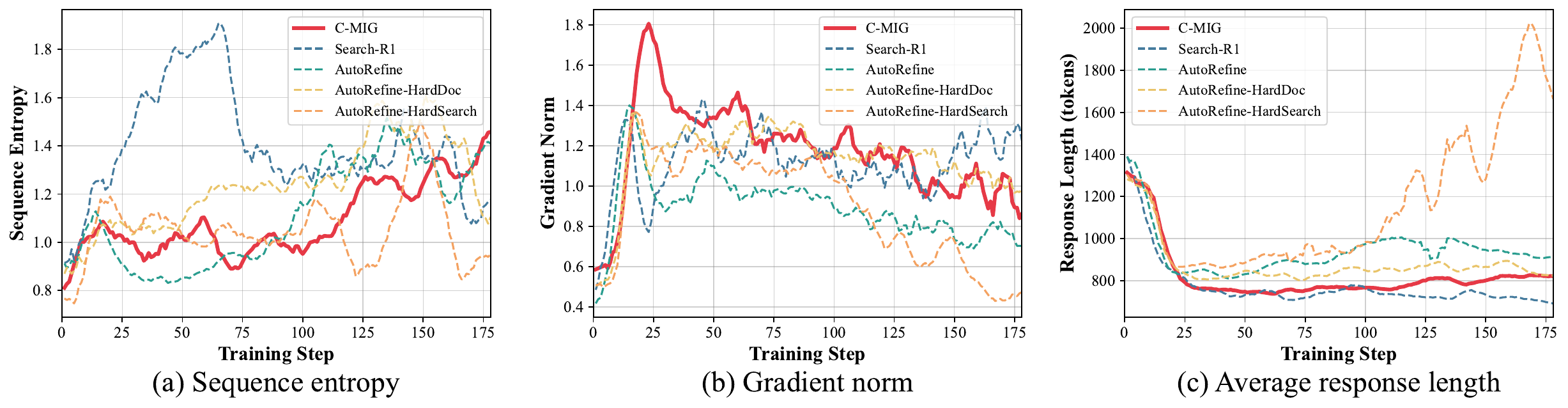}
    \caption{Training dynamics across all methods (after smoothing). (a) Sequence entropy. (b) Gradient norm. (c) Average response length. C-MIG maintains stable entropy and gradient norms while keeping response length comparable to the baselines.}
    \label{fig:app_training_signals}
\end{figure*}

\subsection{Late-Stage Volatility}

We compute the standard deviation of the last 30\% of training steps as a volatility metric, as shown in Table \ref{tab:volatility}.

\begin{table}[h]
\centering
\small
\setlength{\tabcolsep}{4pt}
\begin{tabular}{lcc}
\toprule
Method & Diag.\ $\sigma$ $\downarrow$ & DocHit $\sigma$ $\downarrow$ \\
\midrule
Search-R1 & 0.074 & 0.053 \\
AutoRefine & 0.054 & 0.055 \\
IGPO & 0.071 & 0.063 \\
AutoRefine-HardSearch & 0.067 & 0.112 \\
AutoRefine-HardDoc & 0.070 & 0.060 \\
\textbf{C-MIG} & 0.060 & 0.051 \\
\bottomrule
\end{tabular}
\caption{Late-stage volatility (std of last 30\% steps). C-MIG achieves the lowest combined volatility.}
\label{tab:volatility}
\end{table}

C-MIG exhibits the lowest or near-lowest late-stage volatility in both diagnostic accuracy and document hit rate. Notably, AutoRefine-HardSearch's retrieval volatility (0.112) is 2.2$\times$ that of C-MIG, indicating that hard process rewards in clinical scenarios lead to extremely sparse rewards and consequent training instability.

\subsection{Gradient Norm Stability}

\begin{table}[h]
\centering
\small
\setlength{\tabcolsep}{4pt}
\begin{tabular}{lccc}
\toprule
Method & Mean & Std & CV $\downarrow$ \\
\midrule

Search-R1 & 1.15 & 0.23 & 0.20 \\
AutoRefine & 0.77 & 0.11 & 0.14 \\
IGPO & 0.62 & 0.16 & 0.25 \\
AutoRefine-HardSearch & 0.59 & 0.13 & 0.22 \\
AutoRefine-HardDoc & 1.08 & 0.13 & 0.12 \\
\textbf{C-MIG} & 1.01 & 0.18 & 0.18 \\
\bottomrule
\end{tabular}
\caption{Gradient norm statistics (last 30\% steps). CV: coefficient of variation. IGPO's highest CV reflects gradient noise from reward drift.}
\label{tab:grad_norm}
\end{table}

We compare the gradient norm stability across different models, as illustrated in Figure~\ref{fig:app_training_signals}(b) and Table \ref{tab:grad_norm}.
C-MIG maintains a gradient norm mean of 1.012 (higher than IGPO's 0.622, indicating sufficient learning signal) with a moderate CV of 0.181. IGPO exhibits the highest CV (0.250), confirming that computing information gain with the continuously updated policy model introduces gradient noise due to reward drift.

\begin{table*}[t]
\centering
\renewcommand{\arraystretch}{1.15}
\setlength{\tabcolsep}{4pt}
\resizebox{\textwidth}{!}{%
\begin{tabular}{ll ccc ccc ccc ccc c}
\toprule
\multirow{2}{*}{Backbone} & \multirow{2}{*}{Methods}
 & \multicolumn{3}{c}{MedDDx-Plus}
 & \multicolumn{3}{c}{MedQA}
 & \multicolumn{3}{c}{MedXpertQA}
 & \multicolumn{3}{c}{RJUA}
 & \multirow{2}{*}{\shortstack{Overall\\Avg}} \\
\cmidrule(lr){3-5}\cmidrule(lr){6-8}\cmidrule(lr){9-11}\cmidrule(lr){12-14}
 & & EM & KG & Avg & EM & KG & Avg & EM & KG & Avg & EM & KG & Avg & \\
\midrule
\multirow{4}{*}{Qwen2.5-3B}
  & C-MIG-R w/ Policy  & 48.85 & 55.91 & 52.38          & 5.00  & \textbf{13.10} & 9.05  & 3.96 & 8.20 & 6.08  & 10.43 & \textbf{27.58} & 19.00 & 21.63 \\
    & C-MIG-R w/ Ref     & \textbf{55.03} & \textbf{58.46} & \textbf{56.75} & \textbf{5.50}  & 12.60 & 9.05  & \textbf{4.36} & \textbf{9.70}  & \textbf{7.03}  & \textbf{11.85} & 27.01 & \textbf{19.43} & \textbf{23.07} \\[\tblsep]
    \cdashline{2-15}\noalign{\vskip \tblsep}
  & C-MIG-D w/ Policy  & 54.34 & 57.37 & 55.86          & 5.50  & \textbf{15.30} & \textbf{10.40} & 5.94 & 9.90 & 7.92  & \textbf{11.37} & 24.17 & \textbf{17.77} & 22.99 \\
    & C-MIG-D w/ Ref     & \textbf{56.43} & \textbf{58.66} & \textbf{57.55} & \textbf{7.00}  & 13.20 & 10.10 & \textbf{6.93} & \textbf{11.72} & \textbf{9.32} & 9.00 & \textbf{25.78} & 17.39 & \textbf{23.59} \\
\midrule
\multirow{4}{*}{Qwen2.5-7B}
  & C-MIG-R w/ Policy  & 57.13 & 61.32 & 59.23          & 11.50 & \textbf{24.00} & \textbf{17.75} & 2.77 & 4.91  & 3.84  & 9.95  & 25.12 & 17.54 & 24.59 \\
    & C-MIG-R w/ Ref     & \textbf{60.32} & \textbf{64.87} & \textbf{62.59} & \textbf{12.50} & 21.00 & 16.75 & \textbf{6.34} & \textbf{11.21} & \textbf{8.78}  & \textbf{10.90} & \textbf{27.30} & \textbf{19.10} & \textbf{26.81} \\[\tblsep]
    \cdashline{2-15}\noalign{\vskip \tblsep}
  & C-MIG-D w/ Policy  & 44.37 & 51.74 & 48.05          & \textbf{15.50} & \textbf{25.00} & \textbf{20.25} & \textbf{7.52} & \textbf{12.99} & \textbf{10.25} & 9.95 & 20.19 & 15.07 & 23.41 \\
    & C-MIG-D w/ Ref     & \textbf{53.74} & \textbf{57.73} & \textbf{55.73}          & 11.50 & 21.40 & 16.45 & 6.93 & 12.51 & 9.72  & \textbf{14.69} & \textbf{27.68} & \textbf{21.18} & \textbf{25.77} \\
\bottomrule
\end{tabular}%
}
\caption{Performance comparison of C-MIG variants across in-domain (MedDDx-Plus) and out-of-domain (MedQA, MedXpertQA, RJUA) benchmarks. Across both backbones, scoring with the frozen reference (w/ Ref) consistently outperforms policy self-scoring (w/ Policy).}
\label{tab:cmig-results}
\end{table*}

\begin{table*}[t]
\centering
\renewcommand{\arraystretch}{1.15}
\setlength{\tabcolsep}{4pt}
\resizebox{\textwidth}{!}{%
\begin{tabular}{ll ccc ccc ccc ccc c}
\toprule
\multirow{2}{*}{Backbone} & \multirow{2}{*}{Methods}
 & \multicolumn{3}{c}{MedDDx-Plus}
 & \multicolumn{3}{c}{MedQA}
 & \multicolumn{3}{c}{MedXpertQA}
 & \multicolumn{3}{c}{RJUA}
 & \multirow{2}{*}{\shortstack{Overall\\Avg}} \\
\cmidrule(lr){3-5}\cmidrule(lr){6-8}\cmidrule(lr){9-11}\cmidrule(lr){12-14}
 & & EM & KG & Avg & EM & KG & Avg & EM & KG & Avg & EM & KG & Avg & \\
\midrule
\multirow{4}{*}{Qwen2.5-3B}
  & IGPO w/ Turn      & 51.84 & 54.58 & 53.21          & \textbf{7.00}  & 13.10 & 10.05          & \textbf{5.54} & \textbf{11.01}  & \textbf{8.28}          & 8.53  & \textbf{25.97} & \textbf{17.25} & 22.20 \\
  & IGPO w/ Seq       & \textbf{55.03} & \textbf{58.21} & \textbf{56.62}          & 6.50  & \textbf{14.60} & \textbf{10.55} & 4.16 & 7.17 & 5.66          & \textbf{9.95}  & 23.32 & 16.63 & \textbf{22.37} \\[\tblsep]
   \cdashline{2-15}\noalign{\vskip \tblsep}
  & C-MIG-D w/ Turn   & 50.05 & 52.96 & 51.50          & 6.00  & 12.60 & 9.30          & 6.14 & \textbf{11.49} & 8.82          & \textbf{11.37} & \textbf{27.87} & \textbf{19.62} & 22.31 \\
  & C-MIG-D w/ Seq    & \textbf{56.43} & \textbf{58.66} & \textbf{57.55} & \textbf{7.00}  & \textbf{13.20} & \textbf{10.10}         & \textbf{6.93} & 11.72 & \textbf{9.32} & 9.00  & 25.78 & 17.39 & \textbf{23.59} \\
\midrule
\multirow{4}{*}{Qwen2.5-7B}
  & IGPO w/ Turn      & 52.84 & \textbf{56.21} & \textbf{54.53}          & \textbf{13.00} & 22.60 & \textbf{17.80} & 3.76 & 7.22  & 5.49          & 8.53  & \textbf{28.63} & 18.58 & 24.10 \\
  & IGPO w/ Seq       & \textbf{52.94} & 52.18 & 52.56          & 12.50 & \textbf{22.70} & 17.60         & \textbf{5.74} & \textbf{10.50} & \textbf{8.12}          & \textbf{11.37} & 26.64 & \textbf{19.00} & \textbf{24.32} \\[\tblsep]
   \cdashline{2-15}\noalign{\vskip \tblsep}
  & C-MIG-D w/ Turn   & 0.60  & 10.23 & 5.42           & 0.50  & 2.70  & 1.60          & 0.79 & 4.04  & 2.42          & 0.00  & 0.57  & 0.28          & 2.43 \\
  & C-MIG-D w/ Seq    & \textbf{53.74} & \textbf{57.73} & \textbf{55.73} & \textbf{11.50} & \textbf{21.40} & \textbf{16.45}         & \textbf{6.93} & \textbf{12.51} & \textbf{9.72} & \textbf{14.69} & \textbf{27.68} & \textbf{21.18} & \textbf{25.77} \\
\bottomrule
\end{tabular}%
}
\caption{Comparison of granularity (\textit{w/ Turn} vs.\ \textit{w/ Seq}) for IGPO and C-MIG-D across in-domain (MedDDx-Plus) and out-of-domain (MedQA, MedXpertQA, RJUA) benchmarks. Sequence-level rewards (w/ Seq) consistently outperform turn-level ones (w/ Turn).}
\label{tab:turn-vs-seq}
\end{table*}

\subsection{Response Length}

Figure~\ref{fig:app_training_signals}(c) presents the response length curves.
All methods exhibit a decreasing response length trend (from $\sim$1400 to $\sim$800--1100 tokens), indicating that models learn more concise reasoning expressions. C-MIG's response length remains comparable to other methods, confirming that multi-view information gain rewards do not introduce redundant output.

\section{Case Study}
\label{appendix_case_study}
We present qualitative cases showing how the multi-view information gain in C-MIG enables robust diagnostic reasoning.

\begin{figure*}[t]
    \centering
    \includegraphics[width=\linewidth]{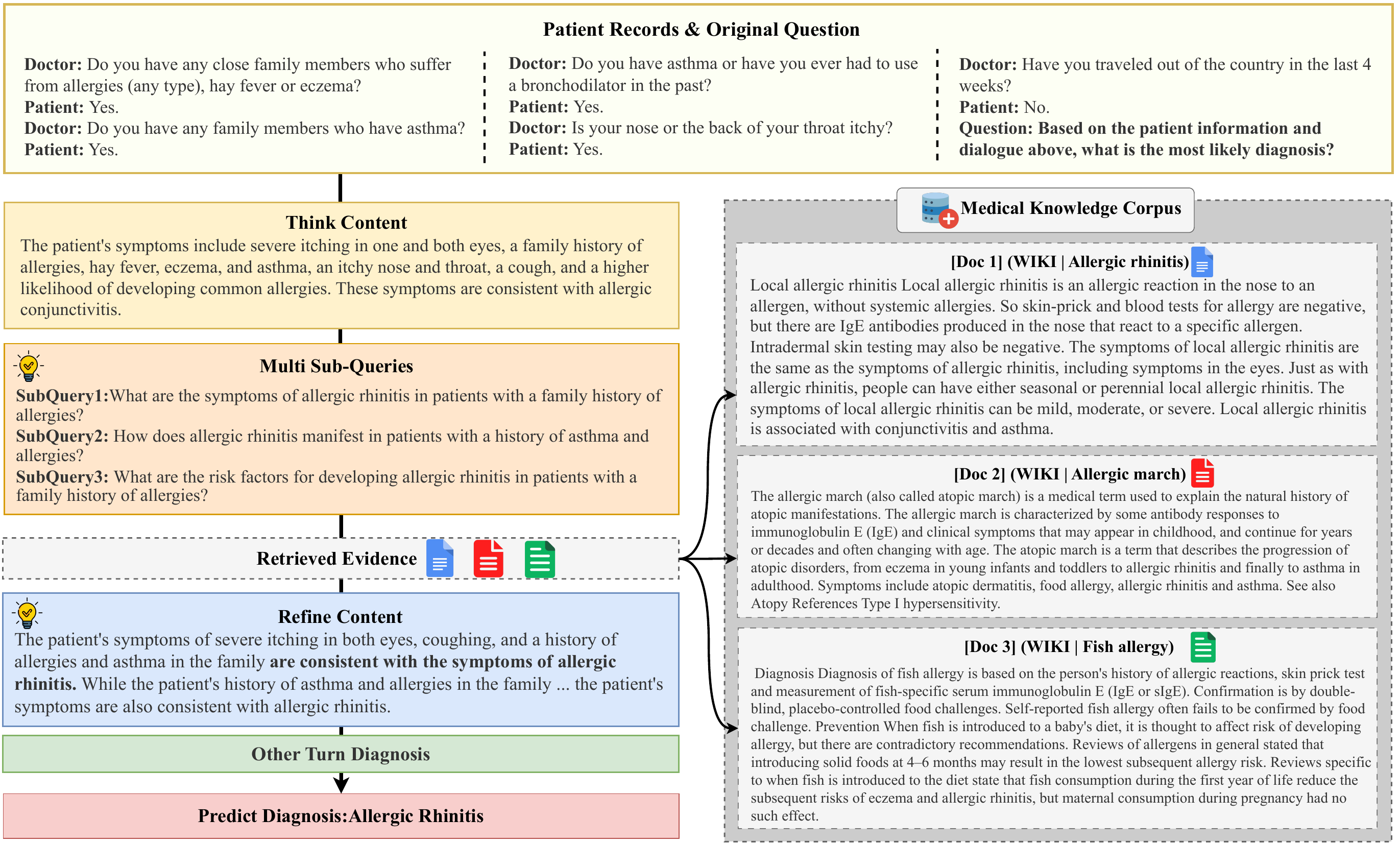}
    \caption{Case study. C-MIG performs step-by-step diagnostic reasoning, crediting semantically relevant evidence that verbatim hard matching would discard.}
    \label{fig:case}
\end{figure*}

\paragraph{Partial Credit under Hard-Reward Failure.}
To make the reward-sparsity problem concrete, we extract from the training logs $7{,}245$ cases in which the hard-match process reward is $0$ yet C-MIG still assigns meaningful partial credit. In the first case the gold diagnosis is anemia: for a 42-year-old male with blood in the stool and a poor diet, the model retrieves a document on fecal occult blood for gastrointestinal-bleeding detection. Hard matching fails because the document discusses fecal occult blood rather than anemia, yet chronic GI blood loss is a major etiology of anemia, and C-MIG detects that the document raises the predicted likelihood of the anemia diagnosis with $\Delta_{\text{IG}}=2.97$, leading to the correct answer. In the second case the gold diagnosis is GERD: the retrieved document discusses hiatal hernia rather than GERD, so hard matching again fails, yet hiatal hernia is a well-known pathological cause of GERD, and C-MIG captures this etiological relevance with $\Delta_{\text{IG}}=2.28$ and reaches the correct diagnosis. These cases concretely show that C-MIG credits clinically relevant evidence that verbatim hard matching discards as zero.

\paragraph{Patient Presentation.}
We further randomly select a sample from the test set, illustrated in Figure~\ref{fig:case}. Given a patient presenting with severe eye itching, family history of allergies/asthma, nasal itching, and cough, the model first summarizes the key symptoms and forms an initial hypothesis of allergic conjunctivitis. The subsequent retrieval returns documents spanning the disease definition of allergic rhinitis, the progression pattern of the allergic march, and the differential factor of fish allergy, providing complementary evidence that a single generic query would be unlikely to cover.

\paragraph{Information Gain-Guided Retrieval and Refinement Quality.}
The retrieved documents contain highly relevant evidence. Under the C-MIG-D reward, the model is incentivized to retrieve documents that maximally increase $\text{LogitP}(a \mid q, D_k)$ relative to the previous turn, effectively filtering noise across rounds. In the refine content, the model synthesizes relevant evidence while discarding irrelevant information such as fish allergy, producing a concise summary: ``the patient's symptoms of severe itching in both eyes, coughing, and a history of allergies and asthma in the family \textbf{are consistent with the symptoms of allergic rhinitis}.'' The C-MIG-R reward ensures that only refinements genuinely increasing diagnostic confidence, those with $\Delta(s_{\text{refine}}) > 0$, receive positive reward, guiding the model to extract diagnostically decisive content rather than passively paraphrasing retrieved documents. The model ultimately arrives at the correct diagnosis of allergic rhinitis through multi-turn reasoning.

\section{Information Gain Analysis of the Reference Model and Policy Model}
\label{appendix_ref_policy}

Complementing the Effectiveness Analysis of the Frozen Reference Model in the main text, this appendix further substantiates that, in clinical diagnosis, a frozen reference model acts as a stable scorer that resists the reward drift of policy self-scoring and thereby provides a more stable training signal. Building upon Qwen2.5-3B and Qwen2.5-7B as backbones, we compare performance under different likelihood-scoring configurations, with full per-dataset results summarized in Table \ref{tab:cmig-results}. We draw the following observations:

(1) w/ Ref consistently outperforms w/ Policy on both the 3B and 7B backbones, which validates that the frozen reference model effectively mitigates reward drift. (2) the gain holds at both scales and is larger on the 7B backbone, $+2.22$ for C-MIG-R and $+2.36$ for C-MIG-D, than on the 3B backbone, $+1.44$ and $+0.60$, indicating that the stabilizing role of the reference model grows with policy capacity.

\section{Analysis of Rollout-Level Reward and Turn-Level Token Reward}
\label{appendix_reward_level}

This appendix analyzes how the process reward should be assigned under medical multi-turn retrieval, comparing dense turn-level token rewards with aggregated rollout-level rewards, and thereby supports the trajectory-level advantage design adopted in C-MIG (Section~\ref{sec:training}). While IGPO employs turn-level information gain as a local reward for tokens within each reasoning turn, C-MIG applies process rewards at the sequence level across entire trajectories. As both compute rewards from inter-turn differentials, we compare whether such signals should be assigned per-turn or aggregated into sequence-level rewards. We evaluate both granularities on IGPO and C-MIG-D using Qwen2.5-3B (Base) and Qwen2.5-7B (Base) as backbones. The experimental results are shown in Table \ref{tab:turn-vs-seq}. Sequence-level rewards yield overall average gains of +0.17 / +1.28 on Qwen2.5-3B and +0.22 / +23.34 on Qwen2.5-7B for IGPO and C-MIG-D respectively, indicating that aggregated trajectory-level signals provide a more stable optimization target than dense turn-level rewards.

Sequence-level approaches consistently outperform turn-level ones, yielding stable improvements particularly on MedDDx. In clinical knowledge bases, the relevance distribution between documents and diagnoses is highly dispersed, such that per-step information gain is both small in magnitude and large in variance. Directly utilizing such high-variance signals as token-level rewards injects noise into advantage estimation. In contrast, sequence-level methods aggregate signals across multiple retrieval rounds into a more stable trajectory-level advantage estimate, exhibiting stronger robustness in medical scenarios.
% (2) Turn-level rewards risk collapse on larger backbones. C-MIG-D w/ Turn degenerates to 2.43 overall average on Qwen2.5-7B, exhibiting clear reward-hacking behavior, whereas its Seq counterpart recovers to the best result of 25.77. This suggests that fine-grained, per-step rewards interact poorly with high-capacity policies under multi-view information-gain signals, and that sequence-level aggregation is essential for stable training in C-MIG.

\begin{table*}[ht]
\centering
\small
\renewcommand{\arraystretch}{1.15}
\setlength{\tabcolsep}{4pt}
\resizebox{\textwidth}{!}{%
\begin{tabular}{l ccc ccc ccc ccc c}
\toprule
\multirow{2}{*}{Methods}
& \multicolumn{3}{c}{MedDDx-Plus}
& \multicolumn{3}{c}{MedQA}
& \multicolumn{3}{c}{MedXpertQA}
& \multicolumn{3}{c}{RJUA}
& \multirow{2}{*}{Overall Avg} \\
\cmidrule(lr){2-4} \cmidrule(lr){5-7} \cmidrule(lr){8-10} \cmidrule(lr){11-13}
& EM & KG & Avg & EM & KG & Avg & EM & KG & Avg & EM & KG & Avg & \\
\midrule
\multicolumn{14}{l}{\textit{Qwen2.5-3B}} \\
AutoRefine-ICDTree     & 43.87 & 64.09 & 53.98 & 3.00 & 12.60 & 7.80 & 4.95 & 9.86 & 7.40 & 9.00 & \textbf{26.36} & \textbf{17.68} & 21.72 \\
+ Doc    & 51.25 & 66.86 & 59.06 & \textbf{5.50} & \textbf{14.20} & \textbf{9.85} & \textbf{7.92} & \textbf{13.07} & \textbf{10.50} & 8.06 & 23.22 & 15.64 & \textbf{23.76} \\
+ Refine & \textbf{52.54} & \textbf{68.85} & \textbf{60.69} & \textbf{5.50} & 13.10 & 9.30 & 3.76 & 7.96 & 5.86 & \textbf{10.90} & 22.27 & 16.59 & 23.11 \\
\midrule
\multicolumn{14}{l}{\textit{Qwen2.5-7B}} \\
AutoRefine-ICDTree     & 47.16 & 60.92 & 54.04 & 11.00 & \textbf{23.70} & \textbf{17.35} & 5.15 & 8.91 & 7.03 & 9.00 & 21.71 & 15.36 & 23.45 \\
+ Doc    & 47.46 & 65.70 & 56.58 & 10.00 & 22.90 & 16.45 & \textbf{6.34} & 11.13 & 8.74 & 16.59 & 25.59 & 21.09 & 25.72 \\
+ Refine & \textbf{55.03} & \textbf{72.64} & \textbf{63.84} & \textbf{11.50} & 21.20 & 16.35 & \textbf{6.34} & \textbf{11.92} & \textbf{9.13} & \textbf{18.01} & \textbf{30.43} & \textbf{24.22} & \textbf{28.39} \\
\bottomrule
\end{tabular}
}
\caption{Scalability of multi-view information gain based on AutoRefine-ICDTree across different models as backbone. Both +Doc and +Refine consistently improve over AutoRefine-ICDTree on both backbones.}
\label{tab:ablation_scal}
\end{table*}

\section{Scalability Analysis of Multi-View Information Gain}
\label{appendix_scalability}

We conduct scalability experiments on multi-view information gain based on Qwen2.5-3B and Qwen2.5-7B, respectively as backbone. The experimental results are shown in Table \ref{tab:ablation_scal}. We can draw the following conclusions:

(1) Across both backbones, both +Doc and +Refine consistently outperform AutoRefine-ICDTree, yielding overall average improvements ranging from $+1.39$ to $+4.94$. This confirms the scalability and effectiveness of our proposed multi-view information gain.
(2) +Refine is preferred on the larger 7B backbone (\textit{e.g.}, $+4.94$ on Qwen2.5-7B), whereas +Doc is more effective on smaller base models ($+2.04$ on Qwen2.5-3B). This indicates that higher-capacity policies benefit more from compressed, semantically-distilled views, while smaller models gain more from the richer signal preserved in raw documents.

\section{Training-Time Computational Overhead Analysis}
\label{appendix_compute_overhead}

The information-theoretic reward signals require additional forward passes through a frozen reference model to compute the information gain $\text{LogitP}(a \mid \mathcal{I}_{\text{extra}}) - \text{LogitP}(a \mid \mathcal{I}_{\text{base}})$ of Eq.~\ref{eq:unified_ig} during the reward assignment phase of each GRPO training step. We employ a persistent multi-GPU worker pool that keeps the Qwen2.5-3B reference model in bfloat16 in GPU memory across all training steps, combined with dynamic token-budget batching that adaptively determines batch sizes from the actual sequence lengths to maximize GPU utilization.

We measure the computational overhead on four H800 GPUs with 80\,GB of memory over 180 training steps. The reference model occupies approximately 6.8 GB per GPU, less than 8.5\% of H800 memory, and co-exists with the actor model without memory contention. Computing the refinement information gain and the per-turn information increase adds 2.81 and 0.33 seconds per step respectively, a total of 3.14 seconds that corresponds to a 3.4\% overhead ratio relative to the total step time. Reward computation itself therefore accounts for less than 5\% of the total training wall-clock, which rises from 3.71 hours to 4.84 hours over 180 steps. The remaining increase is primarily attributable to the longer rollout sequences encouraged by the retrieval reward rather than to the reward computation.

\section{Rollout Prompt}
\label{appendix_prompt}
The prompt template used for rollout generation is given as follows:

\begin{tcolorbox}[promptbox, title=Prompt Template for Rollout Generation]
% \textbf{[System]}\\
% You are an expert medical evaluator. Assess the quality of a candidate
% answer to an open-ended medical question along the following dimensions:
% \textit{factual accuracy}, \textit{clinical relevance}, and \textit{completeness}.

% \medskip
% \textbf{[User]}\\
% Question: \{\{question\}\} \\
% Reference Answer: \{\{reference\}\} \\
% Candidate Answer: \{\{candidate\}\}

% \medskip
% \textbf{[Output Format]}\\
% Return a JSON object with keys \texttt{score} (0--10) and
% \texttt{justification} (a concise explanation in one paragraph).

You are an expert physician skilled in clinical diagnosis with multi-turn search engine calling. 
To answer questions, you must first reason through the available information using \textbf{<think>} and \textbf{</think>}. 
If you identify missing knowledge, you may issue a search request using \textbf{<search>} query \textbf{</search>}. The retrieval system will provide you with the most relevant documents enclosed in \textbf{<evidence>} and \textbf{</evidence>}.
After each search, you need to summarize and refine the existing documents in \textbf{<refine>} and \textbf{</refine>}. 
You may send multiple search requests if needed. 
Once you have sufficient information, provide a concise final answer using \textbf{<diagnosis>} and \textbf{</diagnosis>}. \\

\medskip
Question: \{question\}
\end{tcolorbox}

\end{document}